%% file: main.tex
\documentclass[lettersize,journal]{IEEEtran}

\usepackage[T1]{fontenc}
\usepackage[utf8]{inputenc}
\usepackage{amsmath,amssymb,amsthm}
\usepackage{mathtools}
\usepackage{bm}
\usepackage{graphicx}
\usepackage{booktabs}
\usepackage{multirow}
\usepackage{array}
\usepackage{tabularx}
\usepackage{xcolor}
\usepackage{soul}
\usepackage{url}
\usepackage{hyperref}
\usepackage{cleveref}
\usepackage{algorithm}
\usepackage{algpseudocode}
\usepackage{enumitem}
\usepackage{tikz}
\usetikzlibrary{arrows.meta}
\usepackage{pgfplots}
\pgfplotsset{compat=1.18}
\usepackage{subcaption}
\usepackage{rotating}
\usepackage{makecell}
\usepackage{longtable}
\usepackage{balance}
\usepackage{cite}
\usepackage{adjustbox}
\usepackage[table]{xcolor}

\newcommand{\vpara}[1]{\smallskip\noindent\textbf{#1}}

\definecolor{darkblue}{RGB}{0,51,102}
\definecolor{lightgray}{RGB}{240,240,240}
\definecolor{myred}{RGB}{180,30,30}
\definecolor{mygreen}{RGB}{30,130,30}

\hypersetup{
    colorlinks=true,
    linkcolor=darkblue,
    citecolor=darkblue,
    urlcolor=darkblue,
    pdftitle={Attributed Graph Clustering: A Unified Framework, Comprehensive Review, and Industrial Perspective},
}

\definecolor{colorBest}{RGB}{200, 230, 255}
\definecolor{colorSecond}{RGB}{230, 245, 255}

\newcommand{\res}[2]{#1{\scriptsize$\pm$#2}}

\newcommand{\best}[2]{\cellcolor{colorBest}\textbf{#1}{\scriptsize$\pm$#2}}

\newcommand{\second}[2]{\cellcolor{colorSecond}#1{\scriptsize$\pm$#2}}

\usepackage{mdframed}
\usepackage{fontawesome5}
\usepackage{xcolor}
\newmdenv[
  topline=false,
  bottomline=false,
  rightline=false,
  linewidth=2.5pt,
  linecolor=blue!60!black,
  backgroundcolor=blue!3!white,
  innerleftmargin=8pt,
  innerrightmargin=6pt,
  innertopmargin=4pt,
  innerbottommargin=4pt,
  skipabove=6pt,
  skipbelow=6pt,
]{insightframe}
\newcommand{\insightbox}[2]{%
  \begin{insightframe}
    \noindent\textbf{\textcolor{black!75}{#1}} #2
  \end{insightframe}
}

\begin{document}


\title{Beyond the Academic Monoculture: \\A Unified Framework and Industrial Perspective for\\ Attributed Graph Clustering}

\author{Yunhui~Liu,
        Yue~Liu,
        Yongchao~Liu,
        Tao~Zheng,
        Stan~Z.~Li,~\IEEEmembership{Fellow,~IEEE,}
        Xinwang~Liu,~\IEEEmembership{Senior Member,~IEEE,}
        and Tieke~He,~\IEEEmembership{Member,~IEEE}
\IEEEcompsocitemizethanks{
\IEEEcompsocthanksitem Yunhui~Liu, Tao~Zheng, and Tieke~He are with the State Key Laboratory for Novel Software Technology, Nanjing University, Nanjing, China. E-mail: \{lyhcloudy1225, hetieke\}@gmail.com.
\IEEEcompsocthanksitem Yue~Liu is with the School of Computing, National University of Singapore, Singapore. E-mail: yliu@u.nus.edu.
\IEEEcompsocthanksitem Yongchao~Liu is with Ant Group, Hangzhou, China. 
E-mail: yongchao.ly@antgroup.com.
\IEEEcompsocthanksitem Stan~Z.~Li is with Westlake University, Hangzhou, China. 
E-mail: Stan.ZQ.Li@westlake.edu.cn.
\IEEEcompsocthanksitem Xinwang~Liu is with the College of Computer Science and Technology, National University of Defense Technology, Changsha, China. 
E-mail: xinwangliu@nudt.edu.cn.
}
}


\markboth{Journal of \LaTeX\ Class Files,~Vol.~14, No.~8, August~2021}
{Liu \MakeLowercase{\textit{et al.}}: Attributed Graph Clustering: A Unified Framework, Comprehensive Review, and Industrial Perspective}


\maketitle

\begin{abstract}
Attributed Graph Clustering~(AGC) is a fundamental unsupervised task that partitions nodes into cohesive groups by jointly modeling structural topology and node attributes. 
While the advent of graph neural networks and self-supervised learning has catalyzed a proliferation of AGC methodologies, a widening chasm persists between academic benchmark performance and the stringent demands of real-world industrial deployment. 
To bridge this gap, this survey provides a comprehensive, industrially grounded review of AGC from three complementary perspectives.
First, we introduce the Encode-Cluster-Optimize taxonomic framework, which decomposes the diverse algorithmic landscape into three orthogonal, composable modules: representation encoding, cluster projection, and optimization strategy. This unified paradigm enables principled architectural comparisons and inspires novel methodological combinations.
Second, we critically examine prevailing evaluation protocols to expose the field's \emph{academic monoculture}: a pervasive over-reliance on small, homophilous citation networks, the inadequacy of supervised-only metrics for an inherently unsupervised task, and the chronic neglect of computational scalability. 
In response, we advocate for a holistic evaluation standard that integrates supervised semantic alignment, unsupervised structural integrity, and rigorous efficiency profiling.
Third, we explicitly confront the practical realities of industrial deployment. By analyzing operational constraints such as massive scale, severe heterophily, and tabular feature noise alongside extensive empirical evidence from our companion benchmark, we outline actionable engineering strategies. Furthermore, we chart a clear roadmap for future research, prioritizing heterophily-robust encoders, scalable joint optimization, and unsupervised model selection criteria to meet production-grade requirements.
To support the community and ensure reproducibility, our industrial-scale \emph{benchmark library} and a continuously updated \emph{reading list} are openly available at \url{https://github.com/Cloudy1225/PyAGC}.
\end{abstract}


\section{Introduction}
\label{sec:intro}
\IEEEPARstart{T}{he} ubiquity of graph-structured data, including social networks, e-commerce platforms, knowledge graphs, biological interaction maps, and financial transaction networks, has established the extraction of latent structure from unlabeled graphs as a central challenge in machine learning. 
At the heart of this challenge lies \emph{Attributed Graph Clustering} (AGC), which aims to partition nodes into disjoint, semantically coherent groups by jointly leveraging two sources of information: the \emph{structural topology} encoded in graph edges, and the \emph{attribute semantics} encoded in per-node feature vectors~\cite{PyAGC}. 

Unlike traditional community detection~\cite{Modularity}, which operates solely on graph topology and thus conflates structurally tight but semantically distinct groups, and unlike standard feature clustering methods such as K-Means~\cite{K-Means}, which ignore relational context, AGC must reconcile two fundamentally heterogeneous signals.
This dual requirement makes AGC both scientifically rich and practically indispensable: in industrial settings where ground-truth labels are prohibitively expensive, AGC enables fraud ring detection in transaction networks~\cite{TA-Struc2Vec}, user segmentation for personalized recommendation~\cite{ELCRec}, and community identification in social platforms~\cite{CGC}.

\vpara{The Rapid Evolution of AGC.}
The evolutionary trajectory of AGC can be broadly categorized into three overlapping eras, each characterized by a dominant modeling paradigm. 
The \emph{pre-deep era} (roughly prior to 2016) was dominated by matrix factorization methods~\cite{SA-Cluster} and probabilistic generative models~\cite{BAGC, GBAGC}, demonstrating that topology and attributes must be jointly modeled.
The \emph{early deep era} (2016--2020) was catalyzed by the rise of Graph Neural Networks (GNNs)~\cite{GCN, GNNSurvey1} and variational autoencoders~\cite{GAE, SDCN}, which enabled deep, non-linear representation learning on graph-structured data. 
The \emph{modern era} (2020--present) is defined by the adoption of self-supervised learning paradigms, such as contrastive learning~\cite{NS4GC, MAGI}, feature decorrelation~\cite{CCASSG, DCRN}, and masked autoencoders~\cite{GraphMAE, ProtoMGAE}, alongside a growing recognition that \emph{scalability} is not optional but a prerequisite for real-world deployment~\cite{PyAGC, DinkNet}.

\vpara{The Persistent Academia-Industry Chasm.}
Despite these rapid advances, a significant and widening chasm persists between academic research and industrial reality. The dominant evaluation paradigm exhibits three structural flaws that collectively undermine reproducibility and practical utility:
\begin{enumerate}[leftmargin=*, topsep=2pt]
    \item \textbf{Dataset Myopia.} The vast majority of methods are evaluated almost exclusively on a canonical set of small citation networks (\texttt{Cora}, \texttt{CiteSeer}, \texttt{PubMed})~\cite{DGCSurvey2, Position}, which are characterized by high homophily, textual features, and at most tens of thousands of nodes. Industrial graphs are orders of magnitude larger, frequently heterophilous, and dominated by tabular, heterogeneous features~\cite{GraphLand, PyAGC}.
    
    \item \textbf{Scalability Neglect.} Most state-of-the-art methods rely on full-batch operations, such as computing dense adjacency matrices~\cite{DAEGC, AdaGAE}, performing pairwise similarity computations~\cite{HSAN, NS4GC}, or constructing global subspace representations~\cite{SAGSC, S2CAG}, that scale quadratically or at best linearly but practically fail beyond graphs with $10^5$ nodes due to GPU memory constraints.

    \item \textbf{Metric Paradox.} Although AGC is fundamentally unsupervised, the community has converged on exclusively supervised evaluation metrics (Accuracy, NMI, ARI)~\cite{DGCSurvey1, DGCBench}. 
    This creates a label-fitting incentive: models are tuned to align with human-annotated class labels rather than to discover intrinsic structure. In realistic deployment scenarios where labels are unavailable, such metrics cannot be computed, yet the unsupervised structural quality metrics~\cite{EvalComm} that can be computed (Modularity, Conductance) are systematically underreported~\cite{PyAGC}.
\end{enumerate}

\vpara{Contributions of This Survey.}
This survey provides a comprehensive, structured synthesis of AGC research designed to address these deficiencies. Our contributions are as follows:
\begin{enumerate}[leftmargin=*, topsep=2pt]
\item \noindent\textbf{Unified Taxonomic Framework.} We introduce the \emph{Encode-Cluster-Optimize} (ECO) framework (Section~\ref{sec:eco}), which subsumes all existing methods we are aware of and provides a structured vocabulary for analyzing, comparing, and composing new algorithms.

\item \noindent\textbf{Comprehensive Literature Review.} We provide a structured survey of over 100 AGC methods (Sections~\ref{sec:encode}--\ref{sec:optimize}), organized by the ECO taxonomy. 
We trace the evolutionary trajectory of each module, analyze key methodological transitions, and identify recurring patterns and trade-offs.

\item \noindent\textbf{Holistic Evaluation Analysis.} We analyze evaluation protocols across the literature (Section~\ref{sec:evaluation}), advocate for a holistic protocol mandating both supervised and unsupervised structural metrics alongside efficiency profiling, and synthesize empirical insights (Section~\ref{sec:empirical}) from our companion industrial-scale benchmark~\cite{PyAGC}.

\item \noindent\textbf{Industrial Perspective and Future Roadmap.} We systematically discuss the unique challenges of deploying AGC in production environments (Section~\ref{sec:industrial}), including heterophily, tabular noise, and massive scalability, and chart a clear research roadmap (Section~\ref{sec:future}).
\end{enumerate}

\vpara{Community Resources.} We provide a \href{https://github.com/Cloudy1225/PyAGC/blob/main/AWESOME_AGC.md}{curated AGC paper collection} and an open-source benchmark library, \href{https://github.com/Cloudy1225/PyAGC}{PyAGC}. Further details, including documentation and package installation, are available via our \href{https://pyagc.readthedocs.io/}{ReadTheDocs} and \href{https://pypi.org/project/pyagc}{PyPI} pages.

\vpara{Relationship to Prior Surveys.}
While several existing surveys have reviewed specific aspects of graph clustering, such as clustering paradigms~\cite{DGCSurvey1, DGCSurvey2}, multi-view approaches~\cite{DGCSurvey3}, or temporal dynamics~\cite{BenchTGC}, they either focus on narrow subsets of methods, rely on the flawed academic evaluation paradigm, or lack the industrial perspective. 
Our work distinguishes itself in three key aspects. 
First, ECO provides a strictly more compositional and generative framework than prior taxonomies, and we survey the most recent literature up to 2026, covering significant methodological advances absent from all prior surveys.
Second, we explicitly confront the academia-industry gap, incorporating industrial datasets, unsupervised metrics, and scalability analysis that prior surveys omit. 
Third, our survey is grounded in empirical evidence from our recent PyAGC benchmark~\cite{PyAGC}, which evaluated 17 representative methods across 12 academic and industrial datasets, results we extensively analyze and interpret throughout the survey.

\begin{figure}[t]
  \centering
  \includegraphics[width=\linewidth]{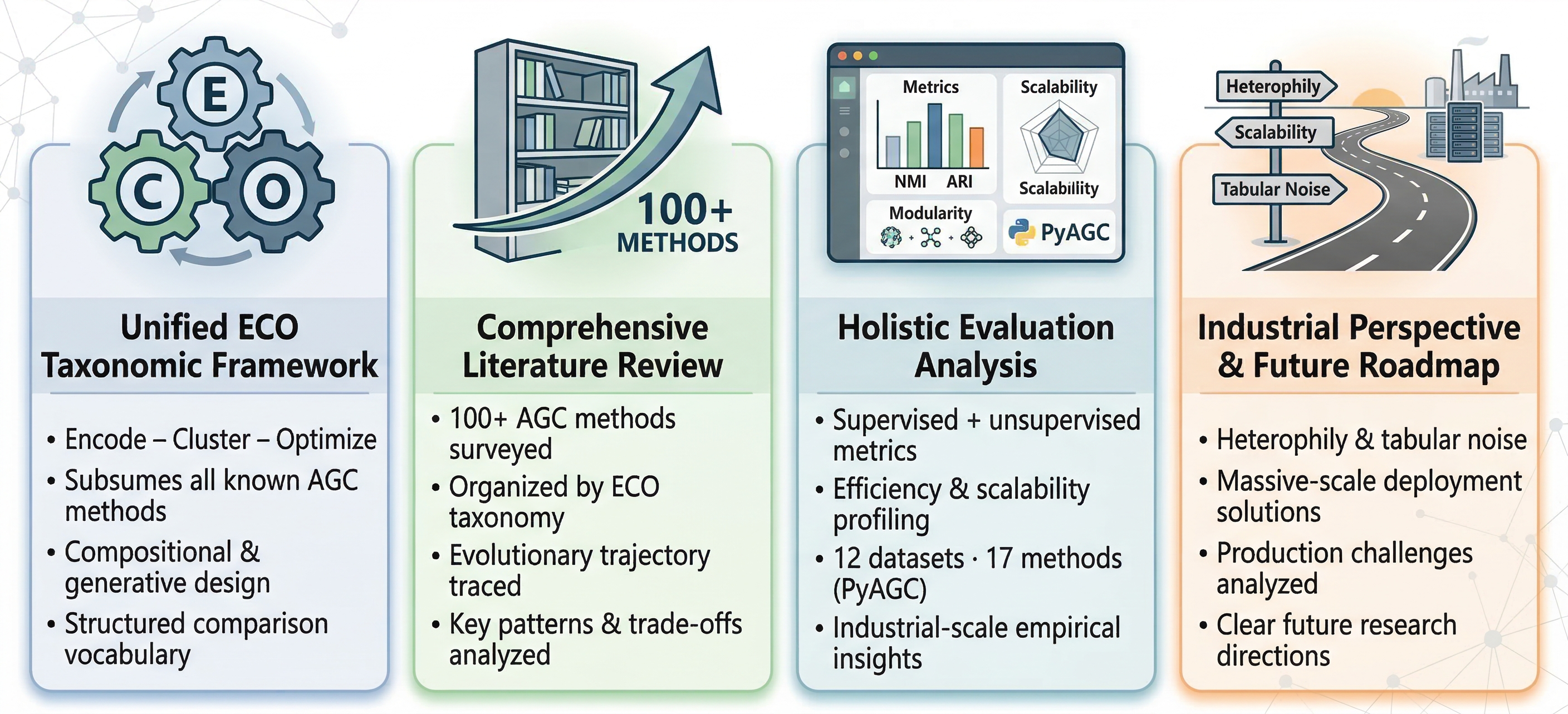}
  \caption{%
    \textbf{Overview of this Survey.}
  }
  \label{fig:contribution}
  \vspace{-5mm}
\end{figure}

\vpara{Relationship to Our Companion Benchmark.}
This survey builds upon and extends our companion work, PyAGC~\cite{PyAGC}, which introduced an industrial-scale benchmark library with standardized implementations and empirical evaluations across 12 datasets and 17 representative methods. Whereas PyAGC focuses on \emph{what} the empirical evidence reveals about current method performance under rigorous conditions, the present survey focuses on \emph{why} the field has evolved as it has, \emph{how} the diverse methodological landscape can be coherently organized, and \emph{where} the most productive research frontiers lie. The empirical findings in Section~\ref{sec:empirical} are drawn from PyAGC and are presented here for analytical completeness; readers seeking full experimental details are referred thereto.

\vpara{Organization.}
The remainder of this survey is organized as follows. 
Section~\ref{sec:prelim} introduces the problem formalization and formally defines the Encode-Cluster-Optimize framework. 
Sections~\ref{sec:encode}, \ref{sec:cluster}, and \ref{sec:optimize} systematically review the existing literature by categorizing methods into the three modules of the ECO framework. 
Section~\ref{sec:extensions} broadens the scope by discussing extensions to complex graph modalities and highlighting downstream applications. 
Section~\ref{sec:evaluation} critically analyzes current evaluation protocols, proposes a holistic evaluation standard, and synthesizes empirical insights from industrial-scale benchmarking. 
Section~\ref{sec:industrial} bridges the academia-industry gap by explicitly detailing the challenges and practical solutions for deploying AGC in real-world production systems. 
Section~\ref{sec:future} charts a comprehensive roadmap for future research directions, and Section~\ref{sec:conclusion} concludes the survey.

\begin{figure*}[t]
\centering
\includegraphics[width=\textwidth]{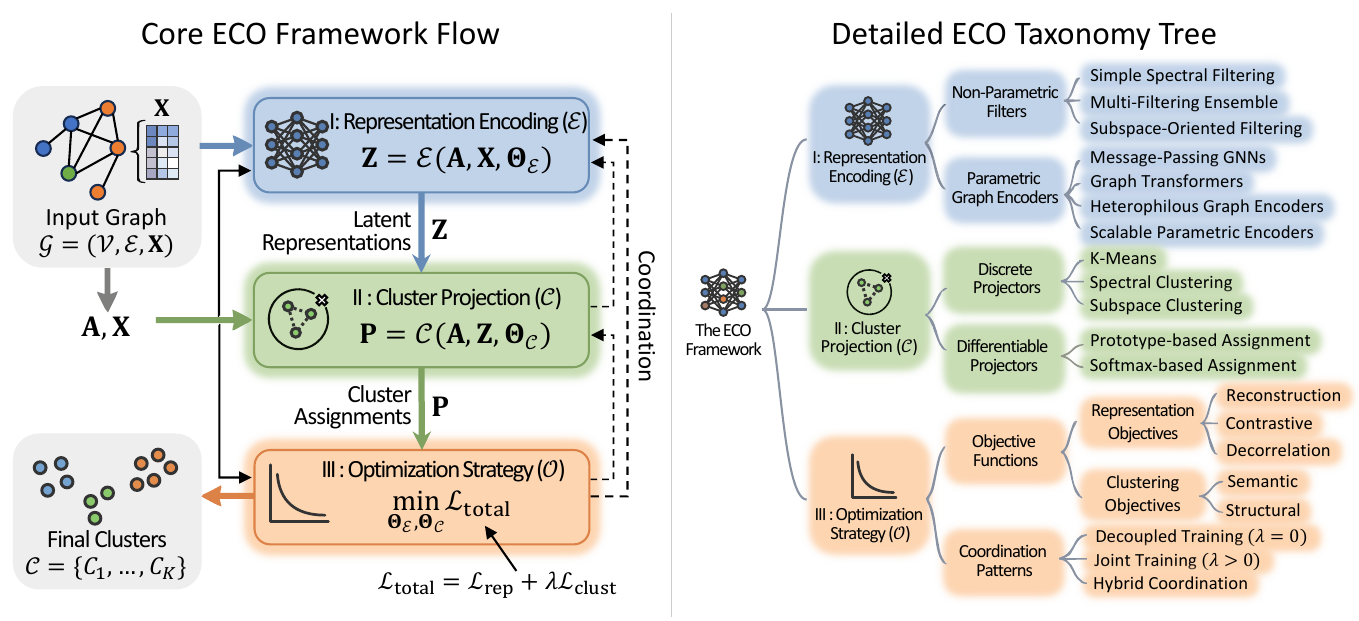}
\caption{\textbf{The Encode-Cluster-Optimize Taxonomic Framework for Attributed Graph Clustering.} Left: Core ECO Framework Flow, showing the sequential data flow, key input/intermediate/output variables, and operational modules. Right: Detailed ECO Taxonomy Tree, providing a hierarchical classification of the methodologies across the three orthogonal modules.}
\label{fig:overview}
\vspace{-5mm}
\end{figure*}

\section{Preliminaries and the ECO Framework}
\label{sec:prelim}
\subsection{Problem Formalization}

\vpara{Attributed Graph.}
An \emph{attributed graph} is a tuple $\mathcal{G} = (\mathcal{V}, \mathcal{E}, \mathbf{X})$, where $\mathcal{V} = \{v_1, \ldots, v_N\}$ is the node set, $\mathcal{E} \subseteq \mathcal{V} \times \mathcal{V}$ is the edge set, and $\mathbf{X} \in \mathbb{R}^{N \times D}$ is the node attribute matrix, with $\mathbf{X}_i \in \mathbb{R}^D$ denoting the feature vector of node $v_i$. The structural topology is encoded by the adjacency matrix $\mathbf{A} \in \{0,1\}^{N \times N}$, where $\mathbf{A}_{ij} = 1$ if $(v_i, v_j) \in \mathcal{E}$.

\vpara{Attributed Graph Clustering.}
Given $\mathcal{G} = (\mathcal{V}, \mathcal{E}, \mathbf{X})$ and a target cluster count $K$, the goal of AGC is to learn a mapping $\mathcal{F}: (\mathbf{A}, \mathbf{X}) \rightarrow \mathbb{R}^{N \times K}$ that assigns each node to one of $K$ disjoint clusters $\mathcal{C} = \{C_1, \dots, C_K\}$, typically represented by a soft assignment matrix $\mathbf{P} \in [0, 1]^{N \times K}$, such that intra-cluster nodes exhibit high structural proximity and attribute similarity, without access to labels.
This distinguishes AGC from: \textit{node classification}, which requires labeled data; \textit{community detection}~\cite{Modularity}, which ignores node attributes; and \textit{generic clustering}~\cite{K-Means}, which ignores graph topology.

\vpara{Homophily and Heterophily.}
A central structural property governing AGC algorithm behavior is \emph{homophily}: the tendency of connected nodes to belong to the same class or cluster. 
Edge homophily $\mathcal{H}_e$ measures the proportion of edges connecting nodes of the same class; node homophily $\mathcal{H}_n$ measures the average fraction of same-class neighbors per node. 
Most AGC methods implicitly assume high homophily ($\mathcal{H}_e \approx 1$); however, industrial graphs (e.g., transaction or web graphs) frequently exhibit strong heterophily ($\mathcal{H}_e < 0.3$), where edges predominantly connect nodes from different clusters. This distinction is a primary driver of the academia-industry performance gap, as we empirically demonstrate in Section~\ref{sec:empirical}.


\subsection{The Encode-Cluster-Optimize (ECO) Framework}
\label{sec:eco}

To impose conceptual order on the diverse AGC landscape, we propose the Encode-Cluster-Optimize framework (Figure~\ref{fig:overview}). ECO decomposes any AGC algorithm into three composable, orthogonal modules:

\begin{equation}
\resizebox{1.0\columnwidth}{!}{%
$\underbrace{\mathbf{Z} = \mathcal{E}(\mathbf{A}, \mathbf{X}; \Theta_{\mathcal{E}})}_{\text{Representation Encoding}} \;\longrightarrow\; \underbrace{\mathbf{P} = \mathcal{C}(\mathbf{A}, \mathbf{Z}; \Theta_{\mathcal{C}})}_{\text{Cluster Projection}} \;\longrightarrow\; \underbrace{\min_{\Theta_{\mathcal{E}}, \Theta_{\mathcal{C}}} \mathcal{L}(\mathbf{A}, \mathbf{X}, \mathbf{Z}, \mathbf{P})}_{\text{Optimization Strategy}}.$%
}
\end{equation}

The three modules are \emph{orthogonal} in the sense that the design choices within each module are largely independent: any encoder $\mathcal{E}$ can in principle be paired with any cluster projector $\mathcal{C}$, governed by any optimization strategy $\mathcal{O}$. They are \emph{composable} in that novel methods arise naturally from combining components across families. We elaborate each module below and in Sections~\ref{sec:encode}--\ref{sec:optimize}.

\vpara{Module I: Representation Encoder ($\mathcal{E}$)}
The encoder $\mathcal{E}$ fuses topological structure and node attributes
into a latent representation $\mathbf{Z} \in \mathbb{R}^{N \times H}$. 
Encoders vary along two principal axes:
(i) \emph{parametric} (GNNs with learnable weights~\cite{GCN,GAT}) versus
\emph{non-parametric} (fixed graph filters~\cite{AGC,SASE,SSGC}); and
(ii) \emph{homophily-based} message passing (which averages neighbor features,
beneficial under high homophily) versus \emph{heterophily-aware} architectures
(which separate ego features from neighborhood features, beneficial under low homophily~\cite{ACMGNN}).

\vpara{Module II: Cluster Projector ($\mathcal{C}$)}
The cluster projector $\mathcal{C}$ maps latent representations to
cluster assignments.
We identify two fundamental paradigms:
(i) \emph{Differentiable projection}, in which the mapping is differentiable, enabling gradient flow from the clustering objective back to the encoder. Instantiations include prototype-based soft assignments~\cite{DAEGC, DinkNet} and differentiable pooling layers~\cite{DMoN, Neuromap}; 
and (ii) \emph{Discrete / Post-hoc projection}, in which a non-differentiable algorithm (e.g., K-Means, Spectral Clustering, Subspace Clustering) is applied to fixed embeddings, decoupling clustering from representation learning~\cite{AGC, SSGC}.
The choice of paradigm has fundamental implications for training stability,
model expressivity, and the risk of degenerate solutions.

\vpara{Module III: Optimization Strategy ($\mathcal{O}$)}
The optimization strategy governs the objective function and the coordination of encoder and clusterer training. The general objective is:
\begin{equation}
  \mathcal{L}_{\text{total}} =
  \underbrace{\mathcal{L}_{\text{rep}}(\mathbf{A}, \mathbf{X}, \mathbf{Z})}_{\text{Representation Loss}}
  + \lambda \underbrace{\mathcal{L}_{\text{clust}}(\mathbf{P}, \mathbf{A})}_{\text{Clustering Loss}},
  \label{eq:total_loss}
\end{equation}
where $\mathcal{L}_{\text{rep}}$ supervises embedding quality~(e.g., via graph reconstruction, contrastive objectives, or feature decorrelation) and $\mathcal{L}_{\text{clust}}$ enforces clustering structure~(e.g., KL divergence, modularity maximization, cut minimization, dilation-shrink loss). 
Their coordination defines three training paradigms:
(i) \emph{Decoupled}: $\mathcal{E}$ is pre-trained with $\mathcal{L}_{\text{rep}}$ alone ($\lambda = 0$), then $\mathcal{C}$ is applied post-hoc;
(ii) \emph{Joint}: $\mathcal{E}$ and $\mathcal{C}$ are optimized simultaneously ($\lambda > 0$), allowing clustering supervision to directly shape the representation space; and
(iii) \emph{Hybrid}: decoupled pre-training is followed by confidence-gated or iterative joint fine-tuning, seeking to inherit the stability of the former and the task-alignment of the latter.

Table~\ref{tab:eco_taxonomy} provides the taxonomy of representative AGC methods, which we discuss in detail throughout Sections~\ref{sec:encode}--\ref{sec:optimize}.

\section{Module I: Representation Encoding}
\label{sec:encode}
The representation encoder aims to produce embeddings that capture both structural proximity and attribute similarity.
We organize existing encoders into two families: non-parametric filters and parametric graph encoders.

\subsection{Non-Parametric Graph Filters}
\label{sec:nonparam}

Non-parametric encoders apply fixed spectral filtering operations to propagate and smooth node attributes over the graph topology, without any learnable weights. Their primary advantages are computational simplicity, absence of gradient instabilities, and theoretical transparency. We categorize them into three families based on their underlying mechanisms.

\vpara{Simple Spectral Filtering.}
The foundation of non-parametric approaches lies in low-pass graph filtering, which smooths node features by aggregating information from neighbors.
The simplest instantiation is SGC~\cite{SGC}, which removes all nonlinearities from GCN and collapses $k$ layers of graph convolution into a single precomputed matrix power:
\begin{equation}
\mathbf{Z} = \tilde{\mathbf{A}}^k \mathbf{X}, \quad \tilde{\mathbf{A}} = \mathbf{D}^{-1/2}(\mathbf{A} + \mathbf{I})\mathbf{D}^{-1/2}.
\end{equation}
SGC achieves remarkable efficiency while retaining competitive performance on homophilous graphs.
A critical limitation of fixed-order smoothing is that the optimal propagation depth $k$ varies drastically across datasets.
AGC~\cite{AGC} addressed this with an adaptive mechanism that iteratively evaluates smoothed representations $\mathbf{Z}^{(k)} = \tilde{\mathbf{A}}^k \mathbf{X}$ and selects the depth $k^*$ minimizing intra-cluster distances, automatically discovering the appropriate smoothness level for diverse graph structures.
IAGC~\cite{IAGC} refined this criterion by jointly considering inconsistencies between filtered features and both the original graph structure and raw features.
SASE~\cite{SASE} further improved scalability by employing random Fourier features to approximate the kernel matrix of smoothed features in linear time.
GRACE~\cite{GRACE} generalized this framework to multiple graph types (undirected, directed, heterogeneous, and hypergraphs) through carefully designed graph Laplacians.

\vpara{Multi-Filtering Ensemble.}
A complementary strategy aggregates features across \emph{multiple} filtering operations, capturing diverse frequency components without explicit order selection.
SSGC~\cite{SSGC} introduced a principled ensemble based on the modified Markov diffusion kernel, computing a convex combination of propagation orders:
\begin{equation}
\mathbf{Z} = \frac{1}{K} \sum_{k=1}^K\left((1-\alpha) \tilde{\mathbf{A}}^k \mathbf{X}+\alpha \mathbf{X}\right),
\end{equation}
where $\alpha$ controls the balance between local ($k=0$) and smoothed ($k \geq 1$) features, yielding a band-pass filter that preserves both high-frequency local variations and low-frequency global patterns.
FGC~\cite{FGC} explicitly constructs representations at multiple filtering depths and learned a weighted combination via a gating network.
NAFS~\cite{NAFS} ensembles fundamentally distinct smoothing operators such as neighbor averaging, degree-normalized averaging, and personalized PageRank diffusion, and notably outperforms many deep GNN competitors on clustering tasks, underscoring the effectiveness of well-designed non-parametric ensembles.

\vpara{Subspace-Oriented Filtering.}
The third family leverages graph filters as a preprocessing step for subspace clustering~\cite{SubCSurvey}, which seeks to express each node embedding as a linear combination of other node embeddings, thereby revealing the latent cluster structure through self-expressiveness.
SAGSC~\cite{SAGSC} applies Laplacian smoothing and feeds the result into a self-expressive module that learns a factored coefficient matrix $\mathbf{Z} = \mathbf{Z}\mathbf{C}$, where $\mathbf{C} \in \mathbb{R}^{N \times N}$ encodes pairwise affinities, with theoretical analysis showing that the smoothing step maximizes the between-to-within cluster distance ratio.
However, explicit construction of the $N \times N$ coefficient matrix incurs quadratic memory cost, limiting scalability. S2CAG and MS2CAG~\cite{S2CAG} resolved this by optimizing a rank-constrained low-rank factorization in $O(N+M)$ time, avoiding materialization of the full coefficient matrix. Notably, these methods establish a theoretical connection between their objective and modularity maximization, demonstrating that subspace clustering implicitly optimizes community structure.

\insightbox{Key Insight.}{
Non-parametric methods offer exceptional computational efficiency, require no training, and avoid the instability associated with gradient optimization. 
They are remarkably effective when the graph exhibits moderate-to-high homophily and when clean textual features are available. 
However, they are inherently limited in their representational capacity: they cannot learn task-specific, non-linear transformations, and their fixed filtering operations may be suboptimal for graphs with complex tabular features or highly heterophilous structure, as we demonstrate in Section~\ref{sec:empirical}.
}

\subsection{Parametric Graph Encoders}
\label{sec:param}

Parametric encoders instantiate $\mathcal{E}$ as a neural network with learnable parameters $\Theta_{\mathcal{E}}$, enabling task-adaptive, non-linear fusion of structure and attributes. Unlike fixed filters, parametric encoders can learn complex, data-driven representations through gradient-based optimization. We organize them into four families based on their architectural design principles.

\vpara{Message-Passing Neural Networks.}
The dominant paradigm for parametric encoding is \emph{message passing}~\cite{GNNSurvey1}, in which each node iteratively aggregates representations from its local neighborhood. 
Graph Convolutional Networks (GCN)~\cite{GCN} instantiate this via a symmetric normalized aggregation:
\begin{equation}
    \mathbf{H}^{(l+1)} = \sigma\!\left(\tilde{\mathbf{A}}\mathbf{H}^{(l)}\mathbf{W}^{(l)}\right),
\end{equation}
where $\tilde{\mathbf{A}}$, is the symmetrically normalized adjacency with self-loops, and $\mathbf{W}^{(l)}$ is a learnable weight matrix. 
The majority of deep AGC methods employ GCNs, including GAE~\cite{GAE}, HSAN~\cite{HSAN}, DinkNet~\cite{DinkNet}, NS4GC~\cite{NS4GC}, and many others.
Graph Attention Networks (GAT)~\cite{GAT} extend GCN by learning adaptive, node-specific aggregation weights, enabling the encoder to selectively attend to
informative neighbors. DAEGC~\cite{DAEGC} employs GAT to capture topological proximity with learnable importance weights. GraphSAGE~\cite{SAGE} further extends message passing to inductive settings via fixed-size neighborhood sampling, enabling mini-batch training~\cite{PyAGC, MAGI}.

\vpara{Graph Transformers.}
Graph Transformers~\cite{GTSurvey} extend the self-attention mechanism of Transformers~\cite{Transformer} to graph-structured data, replacing the local neighborhood constraint with global pairwise attention:
\begin{equation}
    \alpha_{ij} = \frac{\exp\!\left(\mathbf{q}_i^\top \mathbf{k}_j / \sqrt{d} + b_{ij}\right)}{\sum_{k=1}^N \exp\!\left(\mathbf{q}_i^\top \mathbf{k}_k / \sqrt{d} + b_{ik}\right)},
\end{equation}
where $\mathbf{q}_i, \mathbf{k}_j$ are query and key projections of node features, $d$ is the head dimension, and $b_{ij}$ is a structural bias term encoding graph topology (e.g., shortest-path distance)~\cite{Graphormer}. This global receptive field is theoretically more expressive than $L$-hop message passing but incurs $O(N^2)$ complexity, limiting scalability. 
SGFormer~\cite{SGFormer} mitigates this by combining a single-layer global attention term with a local GNN, achieving competitive performance while scaling to 111M-node graphs. Polynormer~\cite{Polynormer} achieves linear complexity by reformulating attention as high-degree equivariant polynomials. 

\vpara{Heterophilous Graph Encoders.}
Standard message-passing encoders implicitly assume homophily: averaging neighbor features is beneficial only when neighbors share the same cluster membership. Under heterophily, such aggregation actively \emph{harms} cluster separability by blending representations across cluster boundaries. ACM-GNN~\cite{ACMGNN} addresses this through adaptive channel mixing that learns per-node convex combinations of low-pass and high-pass filtered representations:
\begin{equation}
\mathbf{H}^{(l+1)}_i = \beta_i (\tilde{\mathbf{A}}\mathbf{H}^{(l)})_i + (1-\beta_i)(\mathbf{H}^{(l)}_i - \tilde{\mathbf{A}}\mathbf{H}^{(l)})_i,
\end{equation}
where $\beta \in [0,1]$ is a learnable node-specific mixing coefficient. 
DGCN~\cite{DGCN} takes a complementary structure-reconstruction approach, constructing separate homophilic and heterophilic graphs and applying a mixed filter to extract both frequency components. 
For AGC, heterophilous encoders are critical for industrial graphs where $\mathcal{H}_e < 0.3$~\cite{PyAGC}, yet remain underexplored relative to their practical importance.

\vpara{Scalable Parametric Encoders.}
Full-batch message passing requires loading the entire graph into GPU memory, incurring $O(N + M)$ complexity that becomes prohibitive for large graphs. Scalable encoders address this via mini-batch training, approximating the full loss over sampled subgraphs $\mathcal{G}_B \sim \mathcal{G}$:
\begin{equation}
    \mathcal{L}_{\text{total}} \approx \mathbb{E}_{\mathcal{G}_B \sim \mathcal{G}}\left[\mathcal{L}(\mathcal{G}_B;\, \Theta_{\mathcal{E}}, \Theta_{\mathcal{C}})\right].
\end{equation}
GraphSAGE~\cite{SAGE} samples fixed-size neighborhoods per hop, reducing memory to $O(Bf^L)$ per mini-batch independent of $N$~\cite{Node2Vec}. 
MAGI~\cite{MAGI} performs two-stage random walk sampling to construct mini-batches with community-aware structure.
ClusterGCN~\cite{ClusterGCN} and GraphSAINT~\cite{GraphSAINT} instead partition the graph into dense subgraphs, providing lower sampling variance. 
As demonstrated in~\cite{PyAGC}, these adaptations enable deep AGC methods to scale to 111M-node graphs on a single GPU, a regime entirely inaccessible to full-batch counterparts.

\insightbox{Key Insight.}{
Parametric encoders offer substantially greater representational flexibility than non-parametric filters, capable of learning complex, non-linear, and task-adaptive feature transformations. 
However, this expressiveness comes with costs: susceptibility to over-smoothing in deep stacks, sensitivity to hyperparameter choices, training instability under joint optimization with clustering objectives, and the fundamental scalability challenge of full-batch processing. 
Heterophilous architectures and Graph Transformers partially address representational limitations, while mini-batch strategies resolve memory bottlenecks at the cost of approximation error.
}

\input{tables/eco_taxonomy}

\section{Module II: Cluster Projection}
\label{sec:cluster}
The cluster projector maps latent embeddings to cluster assignments.
The central design axis is \textit{differentiability}: whether gradients can flow from the clustering step back to the encoder.

\subsection{Discrete Projectors}
\label{sec:discrete}
Discrete projectors apply a traditional, non-differentiable clustering algorithm to fixed encoder outputs, completely decoupling representation learning from cluster assignment.

\vpara{K-Means.}
The majority of deep AGC methods use $k$-means as the post-hoc projector~\cite{DirVAE, S3GC, CoCo}.
While conceptually simple, K-Means imposes a Euclidean, spherical cluster structure that may be ill-suited for complex cluster geometries prevalent in real graphs.
Standard CPU-based K-Means implementations also become prohibitively slow for $N > 10^6$, a bottleneck addressed by PyAGC~\cite{PyAGC} through GPU-accelerated implementations using PyTorch and OpenAI Triton.

\vpara{Spectral Clustering.}
Methods such as AGC~\cite{AGC}, SpectralMix~\cite{SpectralMix}, FGC~\cite{FGC} and SASE~\cite{SASE} use spectral clustering on the smoothed feature similarity matrix, providing a principled non-parametric cluster structure that is more robust to non-spherical geometries.
The main scalability limitation is the $O(N^2)$ cost of constructing the affinity matrix; SASE~\cite{SASE} addresses this using random Fourier feature approximations.

\vpara{Subspace Clustering.}
GALA~\cite{GALA}, DGCSF~\cite{DGCSF}, SAGSC~\cite{SAGSC}, SC3~\cite{SC3}, and S2CAG~\cite{S2CAG} adopt self-expressive subspace clustering, which represents each node as a sparse linear combination of other nodes.
This approach can reveal global cluster structure through the block-diagonal structure of the self-expressive coefficient matrix, and is particularly effective on datasets where clusters correspond to linear subspaces in the feature space.

\subsection{Differentiable Projectors}
\label{sec:differentiable}

Differentiable projectors treat clustering as a trainable layer, enabling the clustering objective to provide supervisory signals that shape the encoder representations.

\vpara{Prototype-Based Assignment.}
Prototype methods maintain a set of $K$ learnable cluster centroids $\{\boldsymbol{\mu}_k\}_{k=1}^K$ and compute soft assignments via a kernel function, commonly the Student's $t$-distribution as introduced by DEC~\cite{DEC}:
\begin{equation}
    \mathbf{P}_{ik} = \frac{(1 + \|\mathbf{z}_i - \boldsymbol{\mu}_k\|^2)^{-1}}{\sum_{k'} (1 + \|\mathbf{z}_i - \boldsymbol{\mu}_{k'}\|^2)^{-1}}.
    \label{eq:student_t}
\end{equation}
DAEGC~\cite{DAEGC}, BTGF~\cite{BTGF}, and GPC~\cite{GPC} apply this in the graph setting, optimizing the KL divergence between the soft assignments $\mathbf{P}$ and a sharpened target distribution.
DinkNet~\cite{DinkNet} reformulates this via Dilation and Shrink loss functions that operate in an adversarial manner to simultaneously encourage cluster compactness and separation, enabling scalable end-to-end training on large graphs.

\vpara{Softmax-Based Direct Assignment.}
DMoN~\cite{DMoN}, Neuromap~\cite{Neuromap}, and DeSE~\cite{DeSE} learn direct assignment matrices $\mathbf{P} \in [0,1]^{N \times K}$ via a softmax-activated MLP head, framing the cluster assignment as a multi-class prediction problem.
The key advantage is that the differentiable assignment matrix can be directly inserted into community-oriented objectives (modularity, cut, map equation), enabling gradient-based optimization of these otherwise discrete measures.

\insightbox{Key Insight.}{
The choice between discrete and differentiable projectors fundamentally trades off \emph{goal alignment} against \emph{training stability}.
Discrete projectors are non-differentiable, preventing clustering supervision from directly shaping embeddings; however, they are theoretically well-understood, computationally efficient, and robust to training instabilities such as representation collapse.
Differentiable projectors enable end-to-end joint optimization, allowing the encoder to be shaped specifically for clustering; however, they risk degenerate solutions (e.g., all nodes assigned to a single cluster)~\cite{DEC} and require careful regularization (e.g., orthogonality constraints)~\cite{MinCut, DMoN}.
Empirically (Section~\ref{sec:empirical}), deep decoupled methods, which use discrete post-hoc projectors, demonstrate greater robustness and stability across diverse datasets and scales.
}

\section{Module III: Optimization Strategy}
\label{sec:optimize}
The optimization strategy specifies the objective function and the interaction between the encoder and the cluster projector. The total loss includes a representation loss and a clustering loss. We survey each component in turn, then analyze the coordination patterns that arise from their interaction.

\subsection{Representation Objectives}
\label{sec:rep_obj}
Representation objectives supervise the quality of the latent embeddings without explicit cluster supervision, drawing on the rich literature of self-supervised learning~\cite{GSSLSurvey1, GSSLSurvey2, GSSLSurvey3}.

\subsubsection{Reconstruction Objectives}
\label{sec:reconstruction}
Reconstruction objectives train the encoder to preserve information from the original graph, operating on either the topology or node attributes.

\vpara{Structure Reconstruction.}
Graph Autoencoders~\cite{GAE} pioneered the use of structure reconstruction for graph representation learning. Given latent embeddings $\mathbf{Z}$, a decoder reconstructs the adjacency matrix via inner products $\hat{\mathbf{A}} = \sigma(\mathbf{Z}\mathbf{Z}^\top)$:
\begin{equation}
    \mathcal{L}_{\text{gae}} = -\sum_{i,j} \left[ \mathbf{A}_{ij} \log \hat{\mathbf{A}}_{ij} + (1-\mathbf{A}_{ij})\log(1-\hat{\mathbf{A}}_{ij}) \right]
\end{equation}
This objective encourages connected nodes to have similar embeddings, which coincides with the clustering objective under homophily. 
The reconstruction paradigm was subsequently adopted in DAEGC~\cite{DAEGC}, ARGA/ARVGA~\cite{ARVGA}, SDCN~\cite{SDCN}, AGCN~\cite{AGCN}, WARGA~\cite{WARGA}, GAE-NA~\cite{GAE-NA} and many joint-training methods. 
A fundamental limitation is \emph{feature drift}: reconstructing the adjacency matrix trains the model to recover all pairwise similarities, including those irrelevant to or detrimental for clustering, causing the latent space to drift away from a clustering-friendly geometry~\cite{RGAE, CVGAE}. 
RGAE~\cite{RGAE} addresses this by introducing a correction operator that gradually transforms the reconstruction target towards a clustering-oriented graph during training. 
Similarly, the dual variational framework of~\cite{BELBO-VGAE} identifies a mismatch between inference and generative models that arises when clustering inductive bias is naively imposed, deriving a tighter variational lower bound that accounts for this discrepancy.

\vpara{Feature Reconstruction.}
GraphMAE~\cite{GraphMAE} shifted the reconstruction target from graph structure to masked node features, borrowing the masked autoencoding paradigm from MAE~\cite{MAE}. A random fraction $\rho$ of node features are masked, and the model is trained to reconstruct them using a cosine-error objective with scaled error:
\begin{equation}
    \mathcal{L}_{\text{feat}} = \frac{1}{|\tilde{\mathcal{V}}|}\sum_{v \in \tilde{\mathcal{V}}} \left(1 - \frac{\mathbf{z}_v^\top \mathbf{x}_v}{\|\mathbf{z}_v\|\|\mathbf{x}_v\|}\right)^\gamma,
\end{equation}
where $\tilde{\mathcal{V}}$ is the masked node set and $\gamma > 0$ is a scaling exponent that focuses training on hard reconstructions. Feature reconstruction is particularly valuable when graph topology is noisy or heterophilous, as it grounds the representations in node attribute space rather than structural proximity. 
ProtoMGAE~\cite{ProtoMGAE} extends this to a multi-granularity setting, combining masked feature reconstruction with prototype-level constraints to produce clustering-friendly embeddings.
A hybrid variant combines both reconstruction targets: DFCN~\cite{DFCN}, EGAE~\cite{EGAE}, CaEGCN~\cite{CaEGCN}, and SAGES~\cite{SAGES} reconstruct both the adjacency and feature matrices using separate decoders, capturing complementary sources of self-supervision.

\subsubsection{Contrastive Objectives}
\label{sec:Contrastive}
Contrastive learning maximizes agreement between representations of semantically related pairs while minimizing agreement with unrelated pairs, providing rich supervisory signals without requiring labels.

\vpara{Augmentation-Based Contrast.}
DGI~\cite{DGI} established the contrastive paradigm for graph learning by maximizing mutual information between local node representations and a global graph summary $\mathbf{s} = \text{readout}(\mathbf{Z})$:
\begin{equation}
    \mathcal{L}_{\text{MI}} = -\frac{1}{N}\sum_{i=1}^N \left[\log \mathcal{D}(\mathbf{z}_i, \mathbf{s}) + \log(1 - \mathcal{D}(\tilde{\mathbf{z}}_i, \mathbf{s}))\right],
\end{equation}
where $\mathcal{D}$ is a discriminator and $\tilde{\mathbf{z}}_i$ are node representations from a corrupted graph. 
MVGRL~\cite{MVGRL} extends this to multi-view contrast across graph diffusion views, while DMGI~\cite{DMGI} adapts the framework to multiplex networks by introducing consensus regularization to align embeddings across different relation types. CommDGI~\cite{CommDGI} further incorporates community structure into the mutual information maximization. GGD~\cite{GGD} introduces a group discrimination paradigm that replaces similarity computation with a simple binary cross-entropy loss, enabling rapid training on billion-edge graphs.

The canonical instance-discrimination framework
, popularized by SimCLR~\cite{SimCLR}, 
defines positive pairs as two augmented views of the same node $(v_i^{(1)}, v_i^{(2)})$ and negatives as all other nodes, optimizing the InfoNCE loss~\cite{InfoNCE}:
\begin{equation}
    \mathcal{L}_{\text{InfoNCE}} = -\frac{1}{N}\sum_{i=1}^N \log \frac{\exp(\mathbf{z}_i^{(1)\top}\mathbf{z}_i^{(2)}/\tau)}{\sum_{j=1}^N \exp(\mathbf{z}_i^{(1)\top}\mathbf{z}_j^{(2)}/\tau)},
\end{equation}
where $\tau$ is a temperature hyperparameter. 
gCooL~\cite{gCooL}, SCGC~\cite{SCGC}, CONGREGATE~\cite{CONGREGATE}, and CONVERT~\cite{CONVERT} all adopt variants of this framework. 
A persistent challenge is the sensitivity of augmentation design: standard augmentations (edge dropping, feature masking) may destroy clustering-relevant structure~\cite{HoLe}, particularly under heterophily.

\vpara{Semantics-Aware Contrast.}
Standard instance-level contrast suffers from \emph{class collision}: nodes from the same cluster are treated as negatives, creating conflicting gradients. 
Several methods address this by incorporating soft semantic information to redefine or reweight contrastive pairs. 
NS4GC~\cite{NS4GC} estimates a reliable node similarity matrix via neighbor alignment and semantic-aware sparsification, defining graded positive/negative relationships beyond binary augmentation pairs. 
HSAN~\cite{HSAN} mines hard positive and negative pairs using both attribute and structure embeddings, dynamically reweighting them under pseudo-label guidance. 
CARL-G~\cite{CARL-G} connects contrastive learning to cluster validation indices, formulating the objective as an internal clustering quality measure that requires neither labels nor augmentations. 
CONGREGATE~\cite{CONGREGATE} performs augmentation-free reweighted contrast in heterogeneous curvature spaces, paying greater attention to hard pairs in the Riemannian product manifold.

\vpara{Structure-Aware Contrast.}
A complementary line of work defines positive and negative pairs directly from graph structural properties (e.g., community membership, random walk proximity, or topological modularity) grounding the contrastive objective in the relational organization of the graph. 
S3GC~\cite{S3GC} uses random walk co-occurrence to define soft positives, preferentially sampling within-cluster pairs under homophily. 
CGC~\cite{CGC} and gCooL~\cite{gCooL} define pairs by hierarchical community membership, jointly learning community partitions and representations. 
MAGI~\cite{MAGI} exploits the modularity matrix $\mathbf{B} = \mathbf{A} - \frac{\mathbf{d}\mathbf{d}^\top}{2|\mathcal{E}|}$ as a natural signed similarity structure: pairs with $\mathbf{B}_{ij} > 0$ serve as positives and $\mathbf{B}_{ij} < 0$ as negatives, providing community-grounded supervision and avoiding augmentation-induced semantic drift.

\vpara{Negative-Free Contrast.}
Negative sampling introduces computational overhead, memory costs, and class collision. BGRL~\cite{BGRL} eliminates negatives through a bootstrapped approach inspired by BYOL~\cite{BYOL}, using an online-target asymmetry to prevent collapse. 
AFGRL~\cite{AFGRL} extends BGRL by discovering nodes that share the local structural information and the global semantics with the graph, while BLNN~\cite{BLNN} explicitly incorporates graph homophily through node-neighbor positive pair expansion, improving intra-class compactness. SSGE~\cite{SSGE} achieves negative-free uniformity by minimizing the KL divergence between learned representations and an isotropic Gaussian, exploiting the fact that Gaussian samples are uniformly distributed on the unit hypersphere. The decorrelation objectives discussed next also discard negative sampling.

\subsubsection{Decorrelation Objectives}
\label{sec:decorrelation}
Decorrelation objectives prevent representation collapse by enforcing statistical independence between feature dimensions, eliminating the need for negative pairs altogether.
CCASSG~\cite{CCASSG} is inspired by Canonical Correlation Analysis and Barlow Twins~\cite{BarlowTwins, GBT}, minimizing the cross-correlation matrix between two augmented views towards the identity:
\begin{equation}
\begin{aligned}
    \mathcal{L}_{\text{BT}} = \underbrace{\sum_{i}(1 - \mathcal{C}_{ii})^2}_{\text{invariance}} + \lambda_{\text{reg}} \underbrace{\sum_{i}\sum_{j \neq i}\mathcal{C}_{ij}^2}_{\text{decorrelation}},
\end{aligned}
\end{equation}
where $\mathcal{C}_{ij} = \frac{\sum_v \mathbf{z}_{v,i}^{(1)}\mathbf{z}_{v,j}^{(2)}}{\sqrt{\sum_v (\mathbf{z}_{v,i}^{(1)})^2}\sqrt{\sum_v (\mathbf{z}_{v,j}^{(2)})^2}}$ defines the normalized cross-correlation matrix, the invariance term pulls the diagonal to 1, and the decorrelation term pushes off-diagonals to 0. The key advantage over contrastive methods is that the loss is computed over the feature dimension $H$ rather than the batch dimension $N$, decoupling computational cost from graph size.

DCRN~\cite{DCRN} and IDCRN~\cite{IDCRN} introduce dual correlation reduction operating on both the sample-level and feature-level correlation matrices simultaneously, reducing redundancy from two perspectives. AGC-DRR~\cite{AGC-DRR} extends this with adversarial augmentation to reduce input-space redundancy, ensuring diversity of augmented view pairs. BTGF~\cite{BTGF} provides a theoretical analysis showing that certain input structures produce negative semi-definite inner products that lower-bound the Barlow Twins loss and proposes a learned graph filter that yields a provably tighter upper bound.

\subsection{Clustering Objectives}
Clustering objectives directly supervise the cluster assignment matrix, encoding domain-specific notions of what constitutes a good clustering. We organize them into two families: semantic objectives that operate in the representation space, and structural objectives that operate with graph topology.

\subsubsection{Semantic Clustering Objectives}
\label{sec:semantic_obj}
Semantic clustering objectives enforce cluster structure in the latent embedding space, typically measuring the alignment or compactness of node embeddings relative to cluster prototypes.

\vpara{KL Divergence Self-Training.} 
DEC~\cite{DEC} introduced the self-training paradigm, in which given soft assignments $\mathbf{P}$ computed via Eq.~\eqref{eq:student_t}, a sharpened target distribution $\mathbf{Q}$ is:
\begin{equation}
    \mathbf{Q}_{ik} = \frac{\mathbf{P}_{ik}^2 / \sum_i \mathbf{P}_{ik}}{\sum_{k'} \mathbf{P}_{ik'}^2 / \sum_i \mathbf{P}_{ik'}}.
\end{equation}
The clustering loss minimizes the KL divergence between the current soft assignment $\mathbf{P}$ and this sharpened target:
\begin{equation}
    \mathcal{L}_{\text{KL}} = \text{KL}(\mathbf{Q} \| \mathbf{P}) = \sum_{i,k} \mathbf{Q}_{ik} \log \frac{\mathbf{Q}_{ik}}{\mathbf{P}_{ik}}.
\end{equation}
The target $\mathbf{Q}$ emphasizes high-confidence assignments and normalizes for cluster size imbalance. This self-training mechanism is widely adopted in graph clustering: DAEGC~\cite{DAEGC}, CaEGCN~\cite{CaEGCN}, SynC~\cite{SynC}, and DyFSS~\cite{DyFSS} all combine it with graph encoders and representation learning losses. 
A known failure mode is \emph{feature randomness}~\cite{RGAE}: noisy early-stage cluster assignments corrupt the target $\mathbf{Q}$, which then reinforces these errors in subsequent iterations. 
RGAE~\cite{RGAE} mitigates this by introducing a sampling operator that gates gradient flow based on assignment confidence, while CDRS~\cite{CDRS} adopts pseudo-label selection strategies that only incorporate high-confidence assignments into the training signal.


\vpara{Dilation and Shrink Loss.}
DinkNet~\cite{DinkNet} proposes a complementary adversarial pair of objectives that explicitly addresses the two competing goals of clustering: separation between clusters and compactness within clusters. Given cluster centers $\{\boldsymbol{\mu}_k\}_{k=1}^K$ and a batch of embeddings $\{\mathbf{z}_i\}_{i=1}^B$, the dilation loss maximizes inter-cluster distances:
\begin{equation}
    \mathcal{L}_{\text{dilation}} = \frac{-1}{(K-1)K} \sum_{i=1}^K \sum_{j \neq i}^K \|\boldsymbol{\mu}_i - \boldsymbol{\mu}_j\|_2^2,
\end{equation}
while the shrink loss minimizes the average distance from all embeddings to all cluster centers:
\begin{equation}
    \mathcal{L}_{\text{shrink}} = \frac{1}{BK} \sum_{i=1}^B \sum_{j=1}^K \|\mathbf{z}_i - \boldsymbol{\mu}_j\|_2^2.
\end{equation}
The shrink loss deliberately pulls embeddings toward \emph{all} centers rather than just the nearest one, avoiding the confirmation bias of KL self-training by preventing early commitment to potentially incorrect assignments. The two losses are optimized in an adversarial manner, creating a dynamic tension that yields well-separated and compact clusters. 

\subsubsection{Structural Clustering Objectives}
\label{sec:structural_obj}
Structural objectives encode community quality directly in terms of graph topology, grounding the clustering in the relational structure of the data rather than geometric properties of the embedding space.

\vpara{Modularity Maximization.}
Modularity~\cite{Modularity} measures the strength of community structure by comparing the density of intra-cluster edges to a null model, as defined in Eq.~\eqref{eq:modularity}. DMoN~\cite{DMoN} introduced a differentiable spectral relaxation of modularity that can be optimized via gradient descent with soft assignments $\mathbf{P}$:
\begin{equation}
        \mathcal{L}_{\text{DMoN}} = -\frac{1}{2|\mathcal{E}|} \text{tr}\!\left(\mathbf{P}^\top \mathbf{B} \mathbf{P}\right) + \frac{\sqrt{K}}{N} \left\| \sum_i \mathbf{P}_i^\top \right\|_F - 1,
\end{equation}
where $\mathbf{B}_{ij} = \mathbf{A}_{ij} - d_i d_j / (2|\mathcal{E}|)$ is the modularity matrix. The first term maximizes modularity while the second enforces balanced cluster sizes. DGCluster~\cite{DGCluster} extends this to a linear-time formulation without a pre-specified $K$; MAGI~\cite{MAGI} uses modularity as a contrastive pretext task rather than a direct optimization target; and MS2CAG~\cite{S2CAG} proposes a subspace clustering method that implicitly maximizes modularity.

\vpara{Cut Minimization.}
The normalized minimum cut problem seeks a partition that minimizes the fraction of edges crossing cluster boundaries~\cite{NormCut}. The continuous relaxation of MinCut~\cite{MinCut} optimizes two complementary terms:
\begin{equation}
    \mathcal{L}_{\text{MinCut}} = -\frac{\mathrm{tr}(\mathbf{P}^\top \mathbf{A}\mathbf{P})}{\mathrm{tr}(\mathbf{P}^\top \mathbf{D}\mathbf{P})} + \left\|\frac{\mathbf{P}^\top\mathbf{P}}{\|\mathbf{P}^\top\mathbf{P}\|_F} - \frac{\mathbf{I}}{\sqrt{K}}\right\|_F,
\end{equation}
where the first term minimizes the normalized cut ratio and the second orthogonality term prevents collapse. NeuroCUT~\cite{NeuroCUT} generalizes this to arbitrary partitioning objectives via reinforcement learning, enabling generalization to unseen partition counts; HoscPool~\cite{HoscPool} extends cut minimization to higher-order motifs by incorporating motif spectral clustering losses.

\vpara{Further Structural Objectives.}
Beyond modularity and cut-based formulations, several additional families of structural objectives have been proposed. The map equation~\cite{MapEq} is an information-theoretic measure based on random walk dynamics that quantifies the minimum description length of a random walker's trajectory; Neuromap~\cite{Neuromap} formulates a fully differentiable tensor version that intrinsically determines the optimal number of modules, penalizing both over- and under-partitioning without requiring $K$ to be pre-specified. Structural entropy~\cite{StruEntropy} provides a complementary information-theoretic criterion: LSEnet~\cite{LSEnet} and DeSE~\cite{DeSE} minimize $\mathcal{H}^{\mathcal{C}}(\mathcal{G}) = -\sum_{k=1}^{K} \frac{g_k}{2|\mathcal{E}|}\log\frac{V_k}{2|\mathcal{E}|}$, where $g_k$ and $V_k$ are the cut size and volume of cluster $C_k$ respectively, encouraging well-connected partitions without the degree-sequence null model assumed by modularity; LSEnet operationalizes this in hyperbolic space via Lorentz-model partitioning trees, DeSE couples it with graph structure learning, and ASIL~\cite{ASIL} establishes a theoretical bound connecting structural entropy to a linear-complexity tree contrastive loss. Finally, GCSBM~\cite{GCSBM} derives loss functions from Stochastic Block Model (SBM)~\cite{SBM} likelihoods, providing a statistically principled, architecture-agnostic objective that jointly captures intra-cluster density and inter-cluster sparsity under the generative model $P(\mathbf{A}|\mathbf{P})$.

\insightbox{Key Insight.}{
Structural clustering objectives encode cluster quality directly in terms of graph topology, and they can be evaluated \emph{without any ground-truth labels}. For example, modularity and conductance scores computed during training provide reliable intrinsic quality signals and can serve as unsupervised model selection criteria~\cite{PyAGC}. Furthermore, Finding 4 in Section~\ref{sec:empirical} reveals informative divergences between structural and semantic metrics: a method may achieve high modularity (finding topologically tight communities) while exhibiting low accuracy against human-annotated labels, indicating that the annotated labels reflect a semantic view that does not coincide with the graph's topological community structure~\cite{PyAGC}. This observation underscores that structural objectives are not merely complementary to semantic ones, which are indispensable for evaluating and optimizing AGC in label-scarce environments.
}

\begin{figure}[t]
    \centering
    \includegraphics[width=1.0\linewidth]{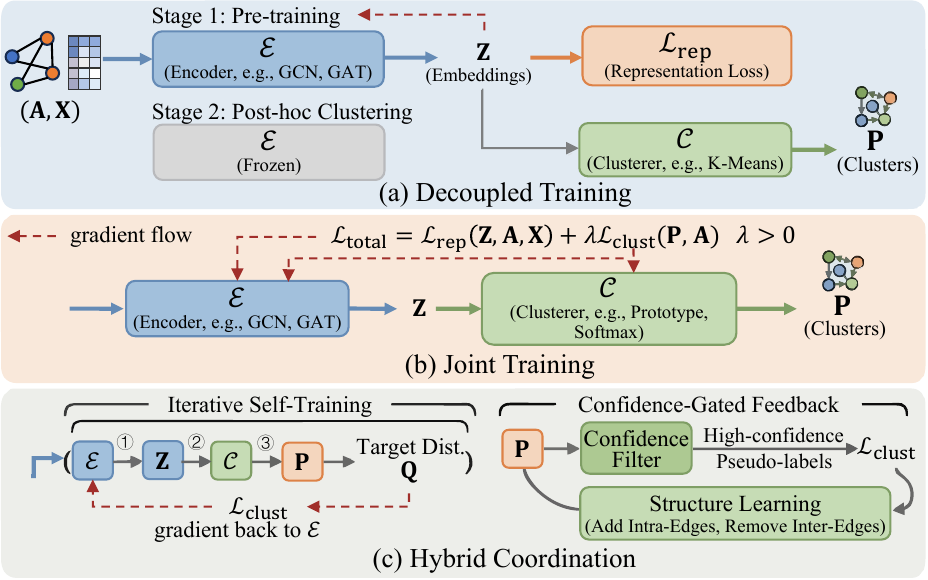}
    \caption{\textbf{Illustration of the three \emph{coordination patterns} in the ECO framework.} \textbf{Decoupled} training applies $\mathcal{L}_\text{rep}$ to pre-train the encoder, then applies the cluster projector post-hoc. \textbf{Joint} training simultaneously optimizes $\mathcal{L}_\text{rep} + \lambda\mathcal{L}_\text{clust}$. \textbf{Hybrid} coordination interleaves these phases via iterative self-training or confidence-gated feedback.}
    \label{fig:coordination}
    \vspace{-5mm}
\end{figure}

\subsection{Coordination Patterns}
\label{sec:coordination}
Beyond the choice of individual objectives, a critical design dimension is how the representation objective and clustering objective are coordinated during training. We identify three principal coordination patterns (Figure~\ref{fig:coordination}), each with distinct implications for stability, performance, and scalability.

\subsubsection{Decoupled Training}
\label{sec:decoupled}

In decoupled training, representation learning and cluster assignment are treated as two entirely separate stages. The encoder is pre-trained using only $\mathcal{L}_{\text{rep}}$ ($\lambda = 0$), after which its weights are frozen and a discrete post-hoc projector is applied to the fixed embeddings. This paradigm spans reconstruction-based methods (GAE~\cite{GAE}, EGAE~\cite{EGAE}), contrastive methods (DGI~\cite{DGI}, NS4GC~\cite{NS4GC}, MAGI~\cite{MAGI}), decorrelation methods (CCASSG~\cite{CCASSG}, DCRN~\cite{DCRN}), and non-parametric filters (SSGC~\cite{SSGC}, SASE~\cite{SASE}), the latter representing the extreme case of eliminating gradient-based encoder training altogether.

\insightbox{Key Insight.}{
The fundamental appeal of decoupled training lies in its separation of concerns: by training $\mathcal{E}$ purely for representation quality, the encoder is insulated from the \emph{feature randomness}~\cite{RGAE} induced by noisy early-stage cluster assignments. The pre-trained encoder also serves as a reusable backbone, and the post-hoc projector can be swapped without retraining. However, this stability comes at the cost of \emph{objective misalignment}: the encoder is optimized for a generic self-supervised proxy only indirectly related to clustering, with no guarantee that the resulting geometry will yield linearly separable clusters. Empirically~\cite{PyAGC}, decoupled methods exhibit strong robustness across diverse datasets but may plateau at suboptimal clustering quality on homophilous benchmarks where task-aligned joint training is feasible.
}

\subsubsection{Joint Training}
\label{sec:joint}
In joint training, both $\mathcal{L}_{\text{rep}}$ and $\mathcal{L}_{\text{clust}}$ are optimized simultaneously ($\lambda > 0$), with gradients from $\mathcal{L}_{\text{clust}}$ directly shaping the encoder parameters $\Theta_\mathcal{E}$. Representative instantiations include DAEGC~\cite{DAEGC} and SDCN~\cite{SDCN} combining reconstruction with KL self-training, DinkNet~\cite{DinkNet} pairing contrastive pre-training with dilation-shrink losses, the differentiable pooling family (MinCut~\cite{MinCut}, DMoN~\cite{DMoN}, Neuromap~\cite{Neuromap}, GCSBM~\cite{GCSBM}) optimizing structural objectives end-to-end, and LSEnet~\cite{LSEnet} and ASIL~\cite{ASIL} jointly minimizing structural entropy with automatic $K$ selection.

\insightbox{Key Insight.}{
Joint training addresses the objective misalignment of decoupled methods by steering the encoder toward geometrically compact intra-cluster and well-separated inter-cluster embeddings, achieving superior results on clean, homophilous datasets~\cite{PyAGC}. However, it introduces well-documented failure modes: \emph{representation collapse}~\cite{DEC, DCRN} when $\mathcal{L}_{\text{clust}}$ overwhelms $\mathcal{L}_{\text{rep}}$; \emph{feature drift}~\cite{RGAE} when reconstruction steers the latent space away from cluster-friendly geometry; and \emph{feature randomness}~\cite{RGAE, CDRS} from corrupted early-stage supervisory signals. The balance coefficient $\lambda$ is a sensitive hyperparameter that varies substantially across settings, and empirically~\cite{PyAGC, DGCBench}, joint methods exhibit higher variance and greater degradation on heterophilous or large-scale graphs where the clustering signal is less reliable.
}

\subsubsection{Hybrid Coordination}
\label{sec:hybrid}

Hybrid coordination patterns interleave decoupled and joint training phases, seeking to inherit the stability of the former and the task-alignment of the latter.

\vpara{Iterative Self-Training.}
The most common hybrid pattern performs \emph{alternating optimization}: cluster assignments from one cycle serve as pseudo-supervision for the encoder in the next. CLEAR~\cite{CLEAR} iteratively refines assignments via optimal transport to suppress noisy inter-class edges. CDRS~\cite{CDRS} promotes only high-confidence assignments as supervisory signals. FastDGC~\cite{FastDGC} performs periodic graph structure updates decoupled from main gradient computation, and MetaGC~\cite{MetaGC} applies meta-learning-based reweighting to down-weight ambiguous edges. CGC~\cite{CGC} and gCooL~\cite{gCooL} jointly learn community partitions and representations within a unified contrastive framework.

\vpara{Confidence-Gated Feedback.}
A refined variant selectively gates supervision based on assignment confidence, implementing a curriculum over $\mathcal{L}_{\text{clust}}$. HSAN~\cite{HSAN} employs dynamic sample weighting guided by high-confidence pseudo-labels. CCGC~\cite{CCGC} constructs contrastive pairs exclusively from high-confidence cluster members, and CONVERT~\cite{CONVERT} generates reliable augmented views via a reversible perturb-recover network. HoLe~\cite{HoLe} activates homophily-enhancing structure learning only when sufficient confidence is available. SynC~\cite{SynC} creates a synergistic encoder-augmentation loop addressing both representation collapse and heterophily, and DeSE~\cite{DeSE} couples graph refinement and clustering through a shared structural entropy objective.

\vpara{Reinforcement-Based Coordination.}
RGC~\cite{RGC} casts cluster number determination as a reinforcement learning problem, using a quality network to evaluate clustering reward for candidate $K$ values. NeuroCUT~\cite{NeuroCUT} generalizes this to arbitrary non-differentiable partitioning objectives, decoupling the parameter space from the partition count and enabling inductive generalization to unseen $K$.

\insightbox{Key Insight.}{
Hybrid coordination is motivated by the observation that neither pure decoupled nor pure joint training is universally superior~\cite{PyAGC, DGCBench, EvalDGC}: decoupled training provides stability but sacrifices task alignment, while joint training provides alignment but risks degenerate solutions under noisy or heterophilous conditions. 
Hybrid methods address these complementary failure modes through controlled feedback loops, confidence-gated supervision, or curriculum-structured transitions, at the cost of additional coordination hyperparameters (e.g., pre-training epochs, confidence thresholds) that interact non-trivially with graph properties. 
}

\section{Extensions and Variants}
\label{sec:extensions}
The core AGC formulation assumes a single homogeneous attributed graph with fully observed node features. We survey the most significant extensions that address richer data modalities and more complex problem settings.

\subsection{Clustering on Complex Graphs}
\vpara{Multi-View and Multi-Modal Graph Clustering.}
Real-world entities are frequently described by multiple, heterogeneous sources of information. In multi-view attributed graph clustering, nodes are associated with $V$ distinct attribute views $\{\mathbf{X}^{(v)}\}_{v=1}^V$ and/or graph topologies $\{\mathbf{A}^{(v)}\}_{v=1}^V$, arising from different types of relationships or modalities~\cite{DGCSurvey3}. 
A central challenge is fusing heterogeneous views. Three principal \emph{fusion strategies} emerge: early fusion~\cite{MAGCN} concatenates features before encoding; late fusion~\cite{CMAGC} encodes views independently and combines at the embedding level; and contrastive cross-view fusion~\cite{MCGC, DMGI} treats different views as positive pairs to distill consistent information. MvAGC~\cite{MvAGC} and MAGC~\cite{MAGC} apply graph filtering per view and learn a consensus graph via anchor-based constrained optimization. 
Beyond standard settings, BMGC~\cite{BMGC} addresses pervasive \emph{view imbalance} through dominant view mining, TOTF~\cite{TOTF} handles \emph{incomplete attributes} via a Train-Once-Then-Freeze framework, and AMMGC~\cite{AMMGC} aligns graph structures across views to mitigate unreliable imputations. For \emph{heterogeneous graphs}, HeCo~\cite{HeCo, HeCo++} proposes co-contrastive learning across network schema and meta-path views, SpectralMix~\cite{SpectralMix} provides joint dimensionality reduction for multi-relational graphs, and VaCA-HINE~\cite{VaCA-HINE} employs variational cluster-aware learning to preserve pairwise proximity and high-order structure. The emergence of foundation model embeddings has further introduced \emph{multimodal attributed graphs}: DMGC~\cite{DMGC} and DGF~\cite{DGF} motivate feature-wise denoising and tri-cross contrastive strategies to address lower cross-modal correlation and intense noise, while DMGC-GTN~\cite{DMGC-GTN} employs a graph transformer to mine complementary cross-modal information.

\vpara{Attributed Hypergraph Clustering.}
Hypergraphs $\mathcal{H} = (\mathcal{V}, \mathcal{F})$ generalize graphs by allowing hyperedges $f \in \mathcal{F}$ to connect arbitrary subsets of nodes, naturally modeling co-authorship, collaborative purchase, and other group interactions. 
AHCKA~\cite{AHCKA} pioneers scalable attributed hypergraph clustering through K-nearest neighbor augmentation and joint hypergraph random walk objectives, with ANCKA~\cite{ANCKA} extending the framework to a unified library supporting simple graphs, hypergraphs, and multiplex graphs. AHRC~\cite{AHRC} eliminates the requirement for pre-specified $K$ through multi-hop modularity maximization. HCN~\cite{HCN} replaces expensive hypergraph convolution with lightweight smoothing preprocessing, with HCN-PAI~\cite{HCN-PAI} additionally addressing partially missing attributes via Dirichlet energy minimization.

\vpara{Dynamic and Temporal Graph Clustering.}
Real-world graphs evolve continuously, requiring methods that track cluster evolution without full recomputation. TGC~\cite{TGC} introduces a general framework for deep temporal graph clustering adapted to the interaction sequence-based batch-processing paradigm, enabling flexible time-space trade-offs. BenchTGC~\cite{BenchTGC} provides the first comprehensive benchmark with standardized frameworks and curated temporal datasets. For dynamic community detection, CGC~\cite{CGC} incorporates incremental learning and change point detection, while RTSC~\cite{RTSC} employs successive snapshot embeddings projected into a shared subspace with low-rank block-diagonal constraints and sparse noise removal.

\vpara{Attribute-Missing Graph Clustering.}
In practical deployments, node attributes are frequently incomplete: users decline to provide profile information, or data collection pipelines suffer partial failures. 
Attribute-missing clustering must jointly handle incomplete features and exploit graph topology for imputation or circumvention. 
CGIR~\cite{CGIR} combines clustering-oriented generative imputation, which constrains the generative adversarial sampling space via subcluster distribution estimates, with an edge attention network that identifies class-specific attributes to avoid redundant reconstruction. DGAC~\cite{DGAC} bypasses imputation entirely by constructing a complementary attribute-affinity graph and exploiting high-order connectivity in both graphs via Dirichlet energy optimization. CMV-ND~\cite{CMV-ND} reframes the problem as multi-view clustering by constructing complementary views from differential neighborhood representations across hop distances.

\input{tables/datasets}

\vpara{Large Language Model-Enhanced Clustering.}
The advent of large language models (LLMs) with powerful semantic understanding capabilities has opened new avenues for text-attributed graph clustering, where node features are natural language descriptions.
GCLR~\cite{GCLR} proposes an active learning framework that selectively queries an LLM oracle for pairwise similarity feedback under a limited budget, incorporating imperfect judgments via noise-controlling fine-tuning. MARK~\cite{MARK} introduces a multi-agent LLM framework where three specialized concept, generation, and inference agents, collaboratively produce ranking-based supervision signals for uncertain boundary nodes identified via perturbation analysis.

\vpara{Federated Graph Clustering.}
Privacy-sensitive applications require distributing computation without centralizing raw data. At the graph level, FedGCN~\cite{FedGCN} extracts prototype features from local cluster structures and uploads condensed signals to a server that generates consensus prototypes via Gaussian estimation. FGCN-DKS~\cite{FGCN-DKS} addresses intra- and inter-client knowledge heterogeneity through dual knowledge separation, retaining personalized variant subgraphs locally while sharing cluster-oriented invariant patterns. At the node level, FedNCN~\cite{FedNCN} restores missing cross-client connections through clustering-guided link mending, using an MLP-based projector for privacy-preserving local representation and a GNN-based generator for global consensus.

\subsection{Downstream Applications}
\label{sec:applications}
AGC serves as a foundational component across diverse downstream pipelines. 
In \emph{fraud detection and anomaly detection}, CARE~\cite{CARE} augments graph adjacency with cluster soft assignments for multi-view anomaly detection, CER-GOD~\cite{CER-GOD} employs clustering-guided edge reweighting to address the homophily trap, and BotHP~\cite{BotHP} identifies spatially dispersed bot collectives through prototype-guided cluster discovery. 
In \emph{recommendation systems}, GraphHash~\cite{GraphHash} reduces embedding table sizes via modularity-based bipartite graph clustering with a provable connection to message-passing. ELCRec~\cite{ELCRec} employs graph clustering for user segmentation in session-based recommendation, discovering behavioral communities that improve personalization. 
In \emph{graph condensation}, ClustGDD~\cite{ClustGDD} synthesizes condensed graphs through homophily-maximizing clustering, and DeepCGC~\cite{DeepCGC} reveals that graph condensation methods converge to class-wise clustering in latent space. 
In \emph{bioinformatics}, GeDGC~\cite{GeDGC} integrates Gaussian mixture models with graph embedding constraints for single-cell multi-omics integration. 
In \emph{medical imaging}, GraphCL~\cite{GraphCL} applies graph clustering to semi-supervised segmentation and DCA~\cite{DCA} produces individualized brain parcellations through graph-guided deep clustering.
In \emph{natural language processing}, Cequel~\cite{Cequel} addresses cost-effective LLM-based text clustering by constructing graph constraints, while TIER~\cite{TIER} leverages similarity-guided contrastive graph clustering to initialize the document taxonomy.

\section{Evaluation Protocols and Empirical Analysis}
\label{sec:evaluation}
Evaluation methodology is a fundamental determinant of which methods are developed, optimized, and ultimately deployed. We critically analyze the dominant evaluation paradigm, advocate for a holistic protocol, and synthesize empirical insights from our benchmark~\cite{PyAGC} that challenge several widely-held assumptions about method effectiveness.

\subsection{Datasets: The ``Cora-fication'' Problem}
\label{sec:datasets}

\begin{figure}[t]
  \centering
  \includegraphics[width=\linewidth]{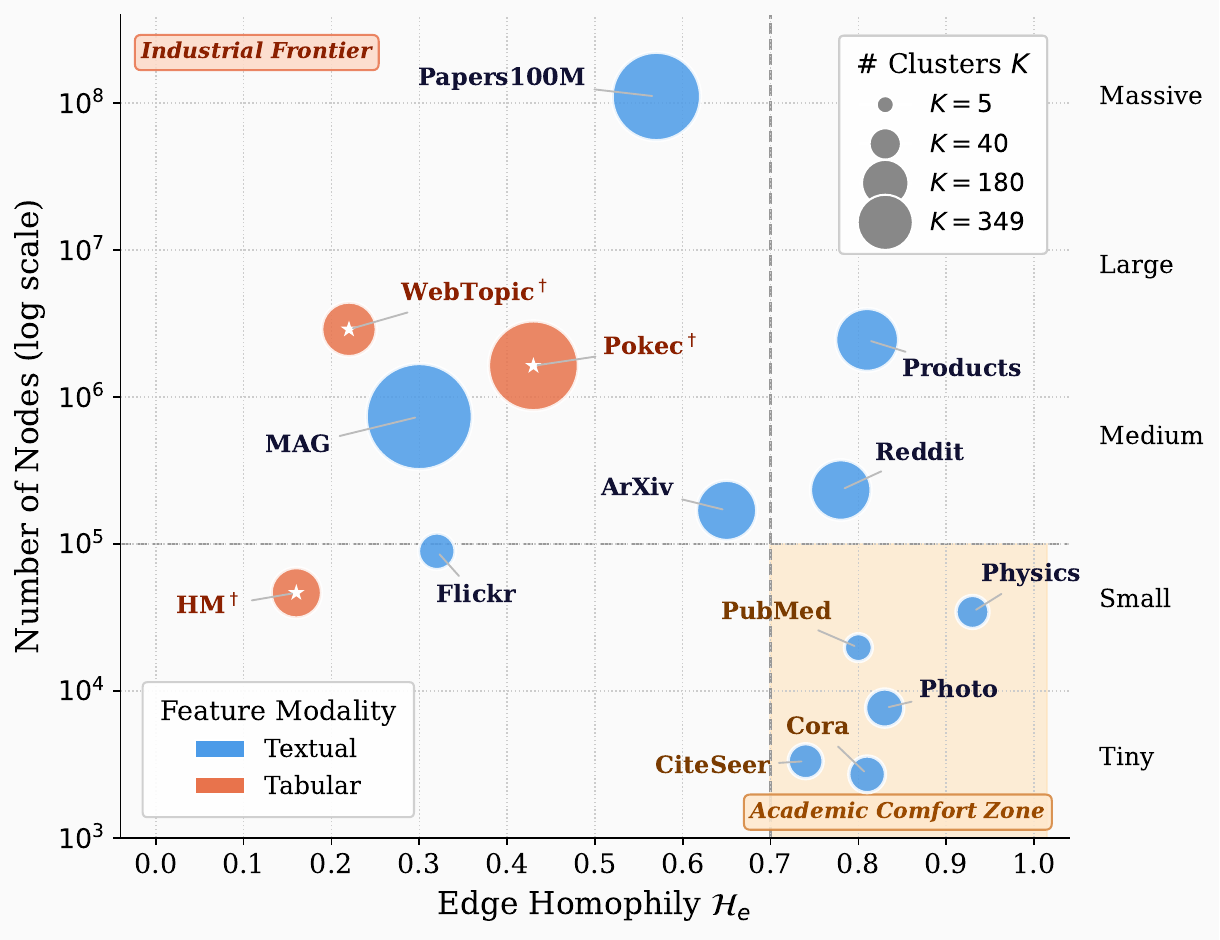}
  \caption{%
    \textbf{Landscape of AGC benchmark datasets} along two axes: edge homophily (\textit{x}-axis) and graph scale (\textit{y}-axis, log scale). Bubble \textbf{color} denotes feature modality and bubble \textbf{size} denotes cluster count~$K$. The shaded region marks the \emph{Academic Monoculture Zone} ($\mathcal{H}_e>0.70$, scale ${<}\,10^5$). Our benchmark~\cite{PyAGC} spans the full landscape across scale, homophily, feature modality, and cluster granularity.
  }
  \label{fig:dataset_landscape}
  \vspace{-5mm}
\end{figure}

\vpara{The Academic Monoculture.}
The vast majority of AGC methods are evaluated on a canonical set of small citation networks: \texttt{Cora}, \texttt{CiteSeer}, and \texttt{PubMed}~\cite{DGCSurvey1, DGCSurvey2, Position}. These datasets share a constellation of properties that render them systematically non-representative of real-world graphs: they contain at most $\sim$20,000 nodes (well within the memory footprint of any GPU), exhibit highs homophily ($\mathcal{H}_e > 0.70$), possess clean bag-of-words textual features, and have cluster counts $K \leq 7$ (Figure~\ref{fig:dataset_landscape}). The consequence of this monoculture is a well-documented publication bias~\cite{Position}: methods are designed to exploit these specific properties, hyperparameters are tuned on these specific datasets, and improvements on these benchmarks are claimed as evidence of general progress.

\begin{figure*}[t]
\centering
\includegraphics[width=1.0\linewidth]{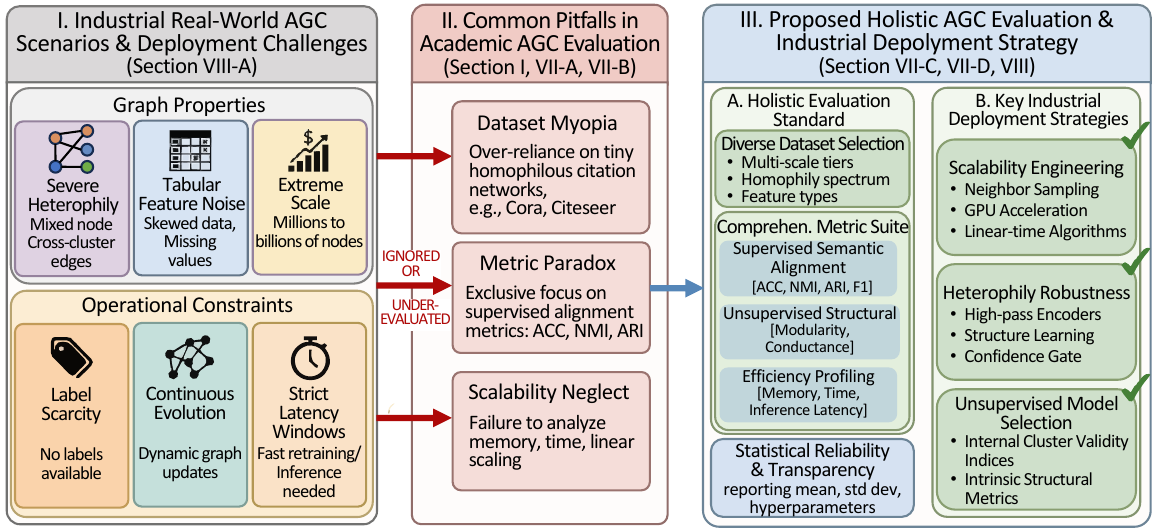}
\caption{\textbf{An \emph{industrial perspective} on attributed graph clustering}, connecting real-world industrial deployment, common academic evaluation pitfalls, and proposed holistic solutions. It aims to bridge the gap between academic research and real-world industrial needs, divided into three interconnected parts: (I) Industrial AGC Scenarios and Deployment Challenges (Sec. \ref{sec:industrial_landscape}), 
(II) Common Pitfalls in Academic AGC Evaluation (Sec. \ref{sec:intro}, \ref{sec:datasets}, \ref{sec:metrics}), (III) Proposed Holistic AGC Evaluation Standard and Key Industrial Deployment Strategies (Sec. \ref{sec:eval_rec} and \ref{sec:scalability_engineering}-\ref{sec:unsupervised_selection})).
}
\label{fig:industrial}
\vspace{-7mm}
\end{figure*}

\vpara{Dimensions of Dataset Diversity.}
A rigorous benchmark must diversify across at least four independent dimensions, as systematized in~\cite{PyAGC}:

\begin{itemize}[leftmargin=*, topsep=2pt]
    \item \textbf{Scale.} Meaningful benchmarks should span multiple orders of magnitude, from thousand nodes to billion nodes of \texttt{Papers100M}~\cite{OGB}, to evaluate both representational quality and scalability under realistic memory constraints.
    
    \item \textbf{Homophily Spectrum.} Including graphs across the full homophily spectrum, from high-homophily citation networks ($\mathcal{H}_e > 0.8$) to high-heterophily web and transaction graphs ($\mathcal{H}_e < 0.25$)~\cite{GraphLand}, is essential for evaluating the robustness of encoders to the fundamental assumption they rely upon.
    
    \item \textbf{Feature Modality.} Industrial graphs are dominated by tabular features: heterogeneous mixtures of categorical indicators, numerical statistics, and behavioral counts~\cite{GraphLand}. Unlike textual bag-of-words features, tabular features exhibit skewed distributions, feature interactions, and domain-specific noise that challenge standard graph encoders.
    
    \item \textbf{Domain and Label Semantics.} Cluster labels in citation networks reflect author-annotated topics with high inter-rater agreement. In industrial graphs, labels may reflect one specific behavioral facet (e.g., product category in \texttt{HM}, geographic region in \texttt{Pokec}) of a multi-faceted entity, resulting in systematic misalignment between topological community structure and semantic labels.
\end{itemize}

Table~\ref{tab:datasets} summarizes the dataset collection of our companion benchmark~\cite{PyAGC}, spanning these four dimensions. We use this collection as the empirical foundation for the analysis below.

\subsection{Metrics: The Supervised Metric Paradox}
\label{sec:metrics}

\subsubsection{Supervised Alignment Metrics}
The community has converged on four supervised metrics measuring the alignment between discovered clusters $\hat{\mathcal{C}}$ and ground-truth labels $\mathcal{Y}$. 

\vpara{Accuracy (ACC) and Macro-F1 (F1).} ACC solves the cluster-to-label assignment via the Hungarian algorithm~\cite{Hungarian} and reports the fraction of correct assignments; Macro-F1 averages per-class F1 scores uniformly, providing balanced evaluation under cluster size imbalance.

\vpara{Normalized Mutual Information (NMI) and Adjusted Rand Index (ARI).} NMI quantifies the shared information between $\hat{\mathcal{C}}$ and $\mathcal{Y}$, normalized to $[0,1]$; ARI measures the fraction of node pairs consistently co-clustered or separated in both partitions, adjusted for chance.
Both are invariant to label permutations.

\insightbox{Critical Limitations.}{
These metrics share a fundamental limitation: they require ground-truth labels, which are unavailable in the label-scarce industrial settings where AGC is most practically valuable. Furthermore, they implicitly assume that the human-annotated labels constitute the \emph{ground truth} for clustering, conflating what is semantically labeled with what is topologically coherent. As demonstrated in~\cite{PyAGC}, these two notions can diverge substantially: a method may discover well-separated topological communities while achieving low ACC because the communities correspond to latent behavioral groupings not captured by the available labels.
Furthermore, these metrics primarily evaluate set-partition alignment rather than topological graph-partitioning quality~\cite{GA-Metric}.
}

\input{tables/nmi_mod}

\subsubsection{Unsupervised Structural Metrics}
Structural metrics evaluate intrinsic partition quality based solely on graph topology, enabling model evaluation and selection without labels.

\vpara{Modularity ($\mathcal{Q}$).}
Modularity~\cite{Modularity} compares intra-cluster edge density to a degree-preserving null model:
\begin{equation}
    \mathcal{Q} = \frac{1}{2|\mathcal{E}|}\sum_{i,j}\left(\mathbf{A}_{ij} - \frac{d_i d_j}{2|\mathcal{E}|}\right)\delta(c_i, c_j),
    \label{eq:modularity}
\end{equation}
where $d_i$ is the degree of node $v_i$ and $\delta(c_i, c_j)$ indicates cluster co-membership. $\mathcal{Q} \in [-1, 1]$ and higher values indicate stronger community structure. It is a widely accepted standard in network science for evaluating community structure.

\vpara{Conductance ($\mathcal{K}$).}
Conductance~\cite{Conductance} measures cluster isolation as the fraction of edge volume crossing cluster boundaries:
\begin{equation}
    \mathcal{K} = \frac{1}{K}\sum_{k=1}^K \frac{c_{C_k}}{2m_{C_k} + c_{C_k}},
\end{equation}
where $c_{C_k}$ is the boundary cut size of cluster $C_k$ and $m_{C_k}$ is the number of internal edges. Lower conductance indicates better-isolated communities. Conductance provides a complementary perspective to modularity: it is sensitive to boundary quality rather than global edge density.

\insightbox{Advocacy for Mandatory Reporting.}{
We advocate strongly for the mandatory reporting of both $\mathcal{Q}$ and $\mathcal{K}$ in all AGC evaluations, even when ground-truth labels are available. The rationale is threefold. First, these metrics enable evaluation on unlabeled industrial graphs. Second, they reveal divergences between topological and semantic quality that supervised metrics obscure. Third, as demonstrated in~\cite{PyAGC}, certain methods that underperform on supervised metrics achieve the highest structural quality, indicating they are more suitable for gang mining and community detection applications where topological coherence matters more than semantic alignment.
}

\subsubsection{Efficiency and Scalability Metrics}

Any evaluation protocol purporting to assess practical utility must include computational profiling:

\begin{itemize}[leftmargin=*, topsep=2pt]
    \item \textbf{Peak GPU Memory Consumption.} The primary scalability constraint for deep methods determines the maximum feasible graph size on a given hardware configuration.
    \item \textbf{Total Training Time.} Encompasses both encoder pre-training and clustering optimization; determines feasibility for high-frequency industrial retraining cycles.
    \item \textbf{Inference Latency.} Time required to produce cluster assignments for new nodes, critical for inductive online deployment scenarios.
    \item \textbf{Scalability Profile.} The empirical relationship between graph size and resource consumption (ideally $O(N)$ or $O(N \log N)$, not $O(N^2)$).
\end{itemize}

\input{tables/efficiency}

\subsection{Empirical Insights from Industrial-Scale Benchmarking}
\label{sec:empirical}

Drawing on the large-scale empirical results of our recent benchmark~\cite{PyAGC} (Tables~\ref{tab:nmi_mod} and~\ref{tab:efficiency}), we distill five empirical findings that challenge or refine conventional wisdom in the AGC literature.
For full details, please kindly read our benchmark paper~\cite{PyAGC} or the supplementary material.

\vpara{Finding 1: The Cora Comfort Zone Does Not Transfer.}
On high-homophily academic datasets, nearly all methods achieve competitive performance: NS4GC achieves 59.40\% NMI on \texttt{Cora}, MS2CAG achieves 72.45\% NMI on \texttt{Physics}, and S3GC achieves 83.45\% NMI on \texttt{Reddit}. However, performance collapses dramatically on industrial graphs. On \texttt{Pokec} ($\mathcal{H}_e = 0.43$, 1.6M nodes, 183 clusters), the best method (SAGSC) achieves only 38.33\% NMI, while most deep joint methods fall below 10\% NMI. On \texttt{HM} ($\mathcal{H}_e = 0.16$), performance is uniformly poor: even the best method (NS4GC, 15.28\% NMI) barely exceeds the KMeans baseline (10.17\% NMI). This collapse reveals that the representational principles enabling success on academic datasets do not generalize to industrial environments with heterophily and tabular noise.

\vpara{Finding 2: Deep Decoupled Methods Exhibit Superior Robustness.}
Across the full dataset collection, Deep Decoupled methods (NS4GC, MAGI, S3GC) demonstrate the most consistent robustness. For example, NS4GC achieves the best NMI on 8 of 12 datasets. In contrast, Deep Joint methods (DAEGC, DinkNet, MinCut, DMoN) exhibit high variance and degraded NMI performance on large-scale and heterophilous graphs (DAEGC collapses to 9.18\% NMI on \texttt{Products} and 28.59\% NMI on \texttt{Papers100M}). This empirically corroborates the concern about feature randomness and training instability in joint training under challenging conditions.

\vpara{Finding 3: Non-Parametric Methods Remain Highly Competitive.}
Despite their architectural simplicity, non-parametric methods consistently outperform many deep learning counterparts across supervised metrics, particularly on high-homophily datasets. MS2CAG achieves 72.45\% NMI on \texttt{Physics}, matching or exceeding all deep methods. SAGSC achieves the best NMI and modularity on \texttt{Pokec} (38.33\% and 50.01\%) and the best modularity on \texttt{Products} (84.56\%) among all methods. However, all three non-parametric methods (SSGC, SAGSC, MS2CAG) incur OOM errors on \texttt{Papers100M} due to full-graph Laplacian operations, whereas mini-batch deep methods (e.g., GAE: 7.58GB, 0.67h; NS4GC: 13.97GB, 1.24h) scale to 111M nodes on a single V100, demonstrating a hard scalability wall that engineering optimization alone cannot surmount.

\vpara{Finding 4: Supervised and Structural Metrics Tell Divergent Stories.}
The divergence between supervised and structural metrics is systematic and informative. On \texttt{Products}, SAGSC achieves the highest modularity (84.56\%), indicating extremely tight communities, yet its NMI (51.78\%) is lower than NS4GC (54.63\%), meaning NS4GC's clusters align better with product category labels while SAGSC discovers denser co-purchase communities more practically relevant for certain downstream tasks (e.g., gang fraud detection). On \texttt{HM}, DMoN achieves the best modularity (12.54\%) despite mediocre NMI (7.59\%), precisely because it directly optimizes a modularity objective rather than label alignment. On \texttt{MAG}, Neuromap achieves the highest modularity (76.25\%) while ranking mid-tier on NMI (39.14\%). These findings empirically validate the \emph{Supervised Metric Paradox} and underscore that structural metrics convey genuinely distinct and valuable information.

\vpara{Finding 5: Scalability Defines the Industrial Boundary.}
The efficiency profiling results (Table~\ref{tab:efficiency}) reveal a clear bifurcation in the AGC landscape along the dimension of scalability, defining a hard boundary between methods that are academically competitive and those that are industrially deployable.
Standard full-batch implementations of DAEGC, NS4GC, and similar methods fail catastrophically on graphs exceeding $\sim$$10^5$ nodes due to GPU memory constraints. The mini-batch implementations in~\cite{PyAGC} effectively decouple memory consumption from graph size. For example, DAEGC trained on \texttt{Papers100M} (111M nodes) consumes only $\sim$16GB on a single V100 (32GB) and completes in 1.4 hours. This demonstrates that deep graph clustering is no longer constrained to toy datasets and is ready for production-grade deployment at the scale of industrial graphs and social networks.

\subsection{Recommendations for Standardized Evaluation}
\label{sec:eval_rec}

Based on the preceding analysis, we propose the following evaluation standard for future AGC research:

\begin{itemize}[leftmargin=*, topsep=2pt]
    \item \textbf{Diverse Dataset Coverage.} 
    Report results on at least one dataset from each scale tier or explicitly acknowledge which scales are out of scope and why. Additionally, include datasets with $\mathcal{H}_e < 0.4$ to evaluate heterophily, which is the dominant condition in industrial deployment.
        
    \item \textbf{Mandatory Structural Metrics.} Report modularity $\mathcal{Q}$ and conductance $\mathcal{K}$ alongside supervised metrics ACC, NMI, ARI, and F1, regardless of label availability.
    
    \item \textbf{Efficiency Profiling.} Report peak GPU memory consumption and total training time, with experiments conducted on a single standardized GPU to enable fair comparison.
    
    \item \textbf{Statistical Reliability.} Report mean and standard deviation over $\ge5$ runs with different random seeds to characterize method variance, particularly important for joint training methods that exhibit high sensitivity to initialization.
    \item \textbf{Hyperparameter Transparency.} Provide all hyperparameters in configuration files or supplementary material, avoiding post-hoc tuning on test sets. The reproducibility crisis in clustering research is substantially driven by undisclosed hyperparameter search over evaluation metrics.

    \item \textbf{Baseline Completeness.} Include at minimum two representatives from each ECO category: non-parametric methods, deep decoupled methods, and deep joint methods, to enable attribution of performance gains to specific design choices rather than category-level advantages.
\end{itemize}

\section{Industrial Challenges and Solutions}
\label{sec:industrial}

The gap between academic attributed graph clustering research and industrial deployment reflects a fundamentally different set of constraints, objectives, and success criteria beyond mere scale differences (Figure~\ref{fig:industrial}).

\subsection{The Industrial Graph Landscape}
\label{sec:industrial_landscape}
Industrial graphs diverge from academic benchmarks along four critical dimensions. 
\emph{Heterophily and complex topology}: transaction networks and web graphs exhibit strong heterophily, where entities deliberately forge cross-cluster connections for camouflage~\cite{BotHP} or naturally bridge categories through complementary purchases~\cite{CoPurchase}. Standard low-pass GNN encoders actively harm performance by blending inter-cluster representations~\cite{ACMGNN, DGCN}. 
\emph{Heterogeneous feature complexity}: industrial node features are typically heterogeneous mixtures of categorical indicators, skewed numerical statistics, and sparse behavioral signals with missing values and temporal distribution shift~\cite{GraphLand}. Standard GNNs, which compute weighted averages of neighbor feature vectors, may be ill-suited when features lack a meaningful Euclidean geometry~\cite{Position}. 
\emph{Extreme scale and efficiency constraints}: production graphs may contain billions of nodes with strict training latency windows (hours, not days), sub-second inference requirements, and constrained serving infrastructure memory. 
\emph{Label scarcity and non-stationarity}: labels are unavailable or available for only a small, potentially biased subset of nodes, their semantics are often ambiguous, and the graph evolves continuously as behaviors change and fraudsters adapt~\cite{TGC}.

\subsection{Scalability Engineering}
\label{sec:scalability_engineering}
Achieving near-linear scalability requires both algorithmic innovation and engineering optimization. For \emph{mini-batch training}, the two principal graph sampling strategies (neighbor sampling~\cite{SAGE, MAGI} and subgraph sampling~\cite{GraphSAINT, ClusterGCN}) trade approximation variance for memory independence from $N$. PyAGC~\cite{PyAGC} demonstrates that these adaptations enable deep AGC on 111M-node graphs within 2 hours on a single V100 GPU. For \emph{clustering primitives}, CPU-based K-Means is fundamentally unscalable beyond $N > 10^6$; GPU-accelerated implementations via PyTorch and OpenAI Triton achieve multi-fold speedups~\cite{PyAGC}, while random Fourier feature approximations~\cite{SASE} enable $O(N)$ spectral clustering. For \emph{algorithmic design}, methods designed with linear-time complexity from the outset, such as ACMin~\cite{ACMin}, MS2CAG~\cite{S2CAG}, SASE~\cite{SASE}, DGCluster~\cite{DGCluster}, achieve scalability gains unreachable by engineering optimization alone. \emph{Distributed training} for billion-edge graphs remains largely unexplored for AGC, as clustering objectives (modularity, KL divergence) require global statistics expensive to synchronize across machines.

\subsection{Handling Heterophily in Production}

Three strategies address heterophily robustness. \emph{Heterophily-aware encoders} directly redesign the message passing mechanism: ACM-GNN~\cite{ACMGNN} adaptively mixes low-pass and high-pass outputs per node, while DGCN~\cite{DGCN} constructs separate homophilic and heterophilic graph views. \emph{Graph structure learning} refines the input graph using cluster assignment confidence as an unsupervised proxy: HoLe~\cite{HoLe} adds intra-cluster edges and removes inter-cluster ones guided by high-confidence assignments, SynC~\cite{SynC} creates a self-reinforcing cycle where refined embeddings guide structure augmentation, and DGAC~\cite{DGAC} bypasses topology entirely by constructing an attribute-affinity graph. \emph{Structural objectives under heterophily}: when edge-density-based metrics (modularity, conductance) become unreliable, structural entropy objectives~\cite{LSEnet, DeSE}, which capture information-theoretic compressibility rather than pure density, provide more robust clustering signals.

\subsection{Unsupervised Model Selection}
\label{sec:unsupervised_selection}
Without ground-truth labels, practitioners cannot compute ACC, NMI, or ARI. Structural metrics (modularity, conductance) provide reliable \emph{relative} rankings among methods on the same dataset~\cite{PyAGC}, supporting multi-objective selection that balances intra-cluster density and inter-cluster separation. Internal cluster validity indices (Silhouette~\cite{Silhouette}, Davies-Bouldin~\cite{DaviesBouldin}) measure embedding-space compactness; CARL-G~\cite{CARL-G} demonstrates that CVI-inspired training objectives~\cite{CVI} simultaneously improve downstream clustering quality. Methods that intrinsically determine $K$, such as map equation~\cite{Neuromap}, structural entropy~\cite{LSEnet, ASIL}, DGCluster~\cite{DGCluster}, are particularly valuable when the cluster count is unknown and may evolve over time.

\subsection{Production-Ready Implementation}

Beyond algorithmic design, reliable deployment requires: \emph{reproducibility} through versioned YAML configuration files that decouple hyperparameters from code~\cite{PyAGC}; \emph{failure mode detection} via automated monitoring of cluster size imbalance, embedding effective rank collapse, and loss divergence; \emph{incremental updating} to avoid full retraining on continuously evolving graphs~\cite{CGC, TGC}; and \emph{interpretability} through structural objective auditing and prototype-based cluster center examination~\cite{DAEGC, DinkNet}.

\section{Future Research Directions}
\label{sec:future}
Drawing from our taxonomy, empirical analysis, and the broader literature, we identify five fundamental open challenges that define the research frontier in AGC.

\vpara{Robust Encoding under Heterophily and Feature Noise.}
The most immediate gap is the lack of AGC methods performing robustly on heterophilous graphs with noisy tabular features~\cite{PyAGC}. Three sub-directions address this: \textbf{(D1.1)} developing principled unsupervised criteria for filter frequency selection under heterophily based on estimated local homophily ratios~\cite{NeuCGC} or spectral properties (without label supervision to guide low-pass vs. high-pass trade-offs); \textbf{(D1.2)} designing feature-aware message passing for tabular graphs that adapts aggregation to feature types (embedding of categoricals, feature-group attention), drawing on tabular deep learning advances~\cite{FT-Transformer, TabPFN}; and \textbf{(D1.3)} developing homophily-agnostic structure learning objectives grounded in attribute affinity rather than structural proximity~\cite{DGAC}, robust to spurious and missing edges in noisy industrial graphs.

\vpara{Unsupervised Model Selection and Evaluation.}
The inability to select models or tune hyperparameters without labels is a critical deployment obstacle. \textbf{(D2.1)} Future work should develop composite unsupervised quality estimators that combine multiple structural signals (modularity, conductance, coverage~\cite{EvalComm}) with provable robustness to the resolution limit~\cite{GA-Metric} and cluster size imbalance. \textbf{(D2.2)} Meta-learning for hyperparameter prediction, e.g., applying graph meta-features to optimal configurations without labels, represents an underexplored direction with high practical impact. \textbf{(D2.3)} Broader community adoption of industrial-scale heterophilous benchmarks requires standardized efficiency profiling protocols and longitudinal performance tracking.

\vpara{Scalable Joint Training.}
The fundamental tension between joint training effectiveness and mini-batch incompatibility demands targeted solutions. \textbf{(D3.1)} Unbiased or low-variance mini-batch estimators of structural objectives (modularity, conductance) are critical: control variate techniques, importance sampling, and local metric approximations~\cite{MAGI} represent promising directions. \textbf{(D3.2)} Principled convergence guarantees for mini-batch prototype synchronization via exponential moving averages or momentum encoders~\cite{MoCo} would advance the theoretical foundations of scalable joint training. \textbf{(D3.3)} Principled curriculum design that schedules the introduction of $\mathcal{L}_{\text{clust}}$ based on measurable assignment confidence properties, rather than heuristic epoch thresholds, would reduce hyperparameter burden and improve cross-dataset generalization.

\vpara{Theoretical Foundations.}
Several fundamental theoretical questions remain open. 
\textbf{(D4.1)} No generalization theory exists for graph clustering analogous to VC-dimension-based bounds for supervised learning; developing such bounds as a function of graph size, homophily, and feature SNR would enable principled method comparison. 
\textbf{(D4.2)} The convergence properties of joint encoder-clusterer training remain poorly characterized even for simplified settings; establishing conditions, particularly for linear encoders on stochastic block model graphs~\cite{SBM}, would provide actionable stabilization insights. 
\textbf{(D4.3)} Deriving bounds on achievable cluster quality as a function of topology, feature informativeness, and target structure, would clarify when AGC is fundamentally feasible and when noise or heterophily precludes reliable clustering. 
\textbf{(D4.4)} A cluster-aware theory of over-smoothing~\cite{OverSmoothing}, quantifying whether intra-cluster representations converge faster than inter-cluster ones, and deriving the minimum useful encoder depth for a given homophily level, would provide principled architectural guidance beyond empirical depth tuning.

\vpara{Foundation Models and Representation Transfer.}
\textbf{(D5.1)} Graph foundation models~\cite{GFMSurvey, TFM4GAD} pre-trained via masked autoencoding or contrastive objectives may encode universal structural priors transferable across domains, but their utility as frozen AGC encoders, and which pre-training objectives produce the most clustering-amenable representations, is largely unexplored. 
\textbf{(D5.2)} Efficient LLM integration for text-attributed graph clustering should move beyond full-graph LLM inference toward distillation into lightweight GNN models, selective querying of boundary nodes, or batched inference strategies that achieve semantic richness at GNN-level scalability. 
\textbf{(D5.3)} Zero-shot graph clustering, such as using prior-data fitted networks~\cite{ZEUS} to generate clustering-friendly embeddings without any graph-specific training, represents a long-horizon vision. 
\textbf{(D5.4)} A principled framework for multimodal industrial graph clustering must handle missing modalities, modality-specific noise levels, and cross-modal semantic alignment, addressing the challenge that different modalities may induce conflicting cluster structures.

\section{Conclusion}
\label{sec:conclusion}
This survey has presented a comprehensive review of attributed graph clustering through the proposed Encode-Cluster-Optimize framework, organizing over 100 methods along representation encoding, cluster projection, and optimization strategy. Additionally, we have critically examined the dominant evaluation paradigm, identifying dataset myopia, scalability neglect, and the supervised metric paradox as impediments to real-world progress. Empirical evidence from our companion PyAGC benchmark reveals that canonical academic benchmarks systematically fail to predict industrial performance. We hope the ECO framework, the holistic evaluation protocol we advocate, and the research roadmap charted across five open frontiers collectively serve as a productive foundation for bridging the persistent gap between academic AGC research and industrial deployment.


\bibliographystyle{IEEEtran}
\bibliography{main}


\vspace{-10mm}
\begin{IEEEbiographynophoto}{Yunhui~Liu}
is a Ph.D. student at Nanjing University. His research interests focus on graph machine learning and structured data analysis.
\end{IEEEbiographynophoto}

\vspace{-10mm}
\begin{IEEEbiographynophoto}{Yue~Liu}
is a Ph.D. student at the National University of Singapore. His research interests focus on self-supervised learning, with a particular emphasis on the safety and intelligence of foundation models, as well as graph learning.
\end{IEEEbiographynophoto}

\vspace{-10mm}
\begin{IEEEbiographynophoto}{Yongchao~Liu}
is a Staff Engineer at Ant Group, focusing on graph learning and agent systems. He previously held research positions at the Georgia Institute of Technology and the University of Mainz.
\end{IEEEbiographynophoto}

\vspace{-10mm}
\begin{IEEEbiographynophoto}{Tao Zheng}
is a Professor at the Software Institute, Nanjing University. His research interests include knowledge graphs and large language models.
\end{IEEEbiographynophoto}

\vspace{-10mm}
\begin{IEEEbiographynophoto}{Stan~Z.~Li} 
is a Professor with Westlake University. He received the Ph.D. degree from Surrey University, UK. His research interests include pattern recognition, machine learning, and AI for Life Science and Drug Design.
\end{IEEEbiographynophoto}

\vspace{-10mm}
\begin{IEEEbiographynophoto}{Xinwang~Liu} 
is a Professor with the College of Computer Science and Technology, National University of Defense Technology. He received the Ph.D. degree from the National University of Defense Technology.
\end{IEEEbiographynophoto}

\vspace{-10mm}
\begin{IEEEbiographynophoto}{Tieke~He}
is an Associate Professor at the Software Institute, Nanjing University. He received the Ph.D. degree from Nanjing University. His research interests include knowledge graphs and large language models.
\end{IEEEbiographynophoto}

\vfill

\end{document}

%% file: tables/eco_taxonomy.tex
\begin{table*}[t]
\centering
\caption{\textbf{Comprehensive ECO Taxonomy of Representative Attributed Graph Clustering Methods.} Complexity denotes the dominant time/space complexity, where $N = |\mathcal{V}|$ and $M = |\mathcal{E}|$.}
\label{tab:eco_taxonomy}
\resizebox{\textwidth}{!}{%
\begin{tabular}{lllllll}
\toprule
\textbf{Method} & \textbf{Venue} & \textbf{Encoder ($\mathcal{E}$)} & \textbf{Cluster ($\mathcal{C}$)} & \textbf{Optimize ($\mathcal{O}$)} & \textbf{Core Objective} & \textbf{Complexity} \\
\midrule
\multicolumn{7}{c}{\textit{\textbf{Non-Parametric \& Decoupled Methods}}} \\
\midrule
AGC~\cite{AGC}       & IJCAI'19  & Simple Filtering & Spectral          & Decoupled & Adaptive Smoothing Order Selection & $O(N^2)$ \\
SSGC~\cite{SSGC}     & ICLR'21   & Multi-Filtering  & K-Means          & Decoupled & Diffusion Kernel Ensemble            & $O(N+M)$ \\
FGC~\cite{FGC}       & SDM'22   & Multi-Filtering  & Spectral          & Decoupled & Multi-hop Filter + Anchor Selection      & $O(N^2)$ \\
GRACE~\cite{GRACE}   & TKDD'22   & Simple Filtering     & K-Means       & Decoupled & Generalized Graph Laplacian Filter   & $O(N+M)$ \\
NAFS~\cite{NAFS}     & ICML'22   & Multi-Filtering    & K-Means          & Decoupled & Multi-Strategy Smoothing Ensemble    & $O(N+M)$ \\
SAGSC~\cite{SAGSC}   & AAAI'23   & Subspace-Oriented   & Subspace         & Decoupled & Self-Expressive Subspace Clustering  & $O(N+M)$ \\
IAGC~\cite{IAGC}     & TKDE'23   & Simple Filtering  & Spectral          & Decoupled & Inconsistency-Aware Adaptive Filter & $O(N^2)$ \\
SASE~\cite{SASE}     & CIKM'24   & Simple Filtering      & Spectral          & Decoupled & Random Fourier Feature Approximation & $O(N+M)$ \\
S2CAG~\cite{S2CAG}   & KDD'25    & Subspace-Oriented      & Subspace  & Decoupled & Rank-Constrained SVD + Conductance    & $O(N+M)$ \\
MS2CAG~\cite{S2CAG}  & KDD'25    & Subspace-Oriented      & Subspace  & Decoupled & Rank-Constrained SVD + Modularity    & $O(N+M)$ \\
CMV-ND~\cite{CMV-ND} & ICML'25   & Multi-Filtering  & K-Means & Decoupled & Complementary Neighborhood Differentiation  & $O(N+M)$ \\
\midrule
\multicolumn{7}{c}{\textit{\textbf{Deep Decoupled Methods}}} \\
\midrule
GAE~\cite{GAE}       & NeurIPS-W'16 & GCN             & K-Means          & Decoupled & Adjacency Reconstruction           & $O(N^2)$ \\
ARGA~\cite{ARVGA}    & IJCAI'18  & GCN           & K-Means  & Decoupled & Reconstruction + Adversarial Reg.    & $O(N^2)$ \\
DGI~\cite{DGI}       & ICLR'19   & GCN             & K-Means          & Decoupled & Mutual Info Maximization (Local-Global)& $O(N+M)$ \\
MVGRL~\cite{MVGRL}   & ICML'20   & GCN          & K-Means          & Decoupled & Multi-View Mutual Info               & $O(N+M)$ \\
CCASSG~\cite{CCASSG} & NeurIPS'21& GCN             & K-Means          & Decoupled & Feature Decorrelation (Barlow Twins) & $O(N+M)$ \\
S3GC~\cite{S3GC}     & NeurIPS'22& GCN             & K-Means          & Decoupled & Scalable Contrastive (Random Walk)            & $O(N^2)$ \\
DCRN~\cite{DCRN}     & AAAI'22   & Mixed GNN   & K-Means          & Decoupled & Dual Correlation Reduction           & $O(N^2)$ \\
BGRL~\cite{BGRL}     & ICLR'22   & GCN             & K-Means          & Decoupled & Bootstrapped Graph Latents           & $O(N+M)$ \\
DGCN~\cite{DGCN}    & ICML'23   & Mixed GNN             & K-Means          & Decoupled & Graph-agnostic Clustering       & $O(N+M)$ \\
NS4GC~\cite{NS4GC}   & TKDE'24   & GCN             & K-Means          & Decoupled & Node Similarity Matrix Guided        & $O(N^2)$ \\
MAGI~\cite{MAGI}     & KDD'24    & GCN / SAGE & K-Means          & Decoupled & Contrastive Modularity               & $O(N^2)$ \\
NeuCGC~\cite{NeuCGC} & TKDE'25   & GCN             & K-Means          & Decoupled & Neutral Contrastive (Homophily-Aware) & $O(N^2)$ \\
CoCo~\cite{CoCo} & ICLR'26   & Mixed GNN             & K-Means          & Decoupled & Compactness and Consistency & $O(N+M)$ \\
\midrule
\multicolumn{7}{c}{\textit{\textbf{Deep Joint Methods}}} \\
\midrule
DAEGC~\cite{DAEGC}   & IJCAI'19  & GAT             & Prototype  & Joint & Reconstruction + KL Divergence       & $O(N^2)$ \\
SDCN~\cite{SDCN}     & WWW'20    & GCN + AE        & Prototype  & Joint & Reconstruction + KL Self-Training    & $O(N^2)$ \\
MinCut~\cite{MinCut} & ICML'20   & GCN             & Softmax        & Joint & Normalized Cut Minimization          & $O(N+M)$ \\
RGAE~\cite{RGAE}     & TKDE'22   & GCN             & Prototype  & Joint & Feature Randomness/Drift Correction  & $O(N^2)$ \\
DMoN~\cite{DMoN}     & JMLR'23   & GCN             & Softmax        & Joint & Spectral Modularity Maximization     & $O(N+M)$ \\
DinkNet~\cite{DinkNet}& ICML'23  & GCN             & Prototype& Joint & Dilation + Shrink Adversarial Loss   & $O(N+M)$ \\
Neuromap~\cite{Neuromap}& NeurIPS'24& GCN           & Softmax        & Joint & Differentiable Map Equation          & $O(N+M)$ \\
DGCluster~\cite{DGCluster}& AAAI'24& GCN            & Softmax        & Joint & Linear-Time Modularity Maximization  & $O(N+M)$ \\
LSEnet~\cite{LSEnet}  & ICML'24  & Lorentz GCN   & Softmax (Tree) & Joint & Structural Entropy (Hyperbolic)        & $O(N^2)$ \\
FVD~\cite{FVD}       & TPAMI'25  & GCN + VAE       & Prototype  & Joint & Wavelet-Enhanced Variational Bound   & $O(N^2)$ \\
ASIL~\cite{ASIL}     & TPAMI'26  & Lorentz GCN   & Softmax (Tree) & Joint & Augmented Structural Entropy Learning  & $O(N^2)$ \\
\midrule
\multicolumn{7}{c}{\textit{\textbf{Hybrid Coordination Methods}}} \\
\midrule
HoLe~\cite{HoLe}     & CIKM'23   & GCN             & K-Means          & Hybrid & Homophily-Enhanced Structure Learning& $O(N^2)$ \\
CLEAR~\cite{CLEAR}   & TIST'23   & GCN             & K-Means     & Hybrid & Optimal Transport + Graph Refinement & $O(N+M)$ \\
HSAN~\cite{HSAN}     & AAAI'23   & GCN             & K-Means          & Hybrid & Hard Sample Aware Contrastive        & $O(N^2)$ \\
CCGC~\cite{CCGC}     & AAAI'23   & GCN      & K-Means          & Hybrid & Cluster-Guided Contrastive           & $O(N^2)$ \\
RGC~\cite{RGC}       & MM'23     & GCN             & K-Means     & Hybrid & Reinforcement Cluster Count Search   & $O(N^2)$ \\
CARL-G~\cite{CARL-G}  & KDD'23 & GCN        & Softmax   & Hybrid & CVI-inspired Representation & $O(N+M)$ \\
NeuroCUT~\cite{NeuroCUT}& KDD'24 & GNN        & Softmax   & Hybrid & Reinforcement-Based Cut Minimization & $O(N+M)$ \\
FastDGC~\cite{FastDGC}& TKDD'24  & GCN      & K-Means          & Hybrid & Periodic Graph Update + Fast Approximation & $O(N^2)$ \\
DGAC~\cite{DGAC}     & WWW'25    & Mixed GNN  & K-Means          & Hybrid & Dirichlet Energy on Dual Graphs      & $O(N^2)$ \\
DeSE~\cite{DeSE}     & KDD'25    & GCN        & Softmax        & Joint & Deep Structural Entropy              & $O(N^2)$ \\
RAGC~\cite{RAGC}     & NeurIPS'25    & Mixed GNN  & K-Means          & Hybrid & Adaptive Augmentation + Contrastive   & $O(N^2)$ \\
\bottomrule
\end{tabular}%
}
\vspace{-5mm}
\end{table*}

%% file: tables/datasets.tex
\begin{table*}[t]
    \centering
    \caption{\textbf{Summary of Representative Datasets for Our Companion Benchmark~\cite{PyAGC}.}
    The collection is categorized by scale, spanning five orders of magnitude from Tiny to Massive. 
    The datasets cover a wide range of domains and feature types (Textual, Tabular).
    $\mathcal{H}_e$ and $\mathcal{H}_n$ denote edge homophily and node homophily, respectively.
    Datasets marked with $\dagger$ are \textbf{industrial graphs} with complex tabular features and low homophily, representing the industrial evaluation frontier~\cite{GraphLand}.
    }
    \label{tab:datasets}
    \resizebox{\textwidth}{!}{
    \begin{tabular}{lll rr rr c c c c}
        \toprule
        \textbf{Scale} & \textbf{Dataset} & \textbf{Domain} & \textbf{\#Nodes} & \textbf{\#Edges} & \textbf{Avg. Deg.} & \textbf{\#Feat.} & \textbf{Feat. Type} & \textbf{\#Clus.} & \textbf{$\mathcal{H}_e$} & \textbf{$\mathcal{H}_n$} \\
        \midrule
        \multirow{2}{*}{Tiny} 
          & Cora & Citation & 2,708 & 10,556 & 3.9 & 1,433 & Textual & 7 & 0.81 & 0.83 \\
          & Photo & Co-purchase & 7,650 & 238,162 & 31.1 & 745 & Textual & 8 & 0.83 & 0.84 \\
        \midrule
        \multirow{3}{*}{Small} 
          & Physics & Co-author & 34,493 & 495,924 & 14.4 & 8,415 & Textual & 5 & 0.93 & 0.92 \\
          & HM$^\dagger$ & Co-purchase & 46,563 & 21,461,990 & 460.9 & 120 & Tabular & 21 & 0.16 & 0.35 \\
          & Flickr & Social & 89,250 & 899,756 & 10.1 & 500 & Textual & 7 & 0.32 & 0.32 \\
        \midrule
        \multirow{3}{*}{Medium} 
          & ArXiv & Citation & 169,343 & 1,166,243 & 6.9 & 128 & Textual & 40 & 0.65 & 0.64 \\
          & Reddit & Social & 232,965 & 23,213,838 & 99.6 & 602 & Textual & 41 & 0.78 & 0.81 \\
          & MAG & Citation & 736,389 & 10,792,672 & 14.7 & 128 & Textual & 349 & 0.30 & 0.31 \\
        \midrule
        \multirow{3}{*}{Large} 
          & Pokec$^\dagger$ & Social & 1,632,803 & 44,603,928 & 27.3 & 56 & Tabular & 183 & 0.43 & 0.39 \\
          & Products & Co-purchase & 2,449,029 & 61,859,140 & 25.4 & 100 & Textual & 47 & 0.81 & 0.82 \\
          & WebTopic$^\dagger$ & Web & 2,890,331 & 24,754,822 & 8.6 & 528 & Tabular & 28 & 0.22 & 0.18 \\
        \midrule
        Massive & Papers100M & Citation & 111,059,956 & 1,615,685,872 & 14.5 & 128 & Textual & 172 & 0.57 & 0.50 \\
        \bottomrule
    \end{tabular}%
    }
    \vspace{-5mm}
\end{table*}

%% file: tables/nmi_mod.tex
\begin{table*}[t]
\centering
\caption{
\textbf{Clustering performance comparison measured by NMI and Modularity (\%) (Mean $\pm$ SD).}
The {\textbf{best}} and \underline{second-best} results are highlighted.
``--'' denotes OOM errors on a single NVIDIA V100 GPU (32GB) as these methods strictly require full-graph processing.
Results are drawn from our companion benchmark~\cite{PyAGC}.}
\label{tab:nmi_mod}
\begin{adjustbox}{width=\textwidth, center}
\begin{tabular}{llll cccccccccc cc}
\toprule
~ & \multirow{2}{*}{\textbf{Model}} & \multirow{2}{*}{\textbf{Metric}} & \multicolumn{2}{c}{\textbf{Tiny}} & \multicolumn{3}{c}{\textbf{Small}} & \multicolumn{3}{c}{\textbf{Medium}} & \multicolumn{3}{c}{\textbf{Large}} & \multicolumn{1}{c}{\textbf{Massive}} \\
\cmidrule(lr){4-5} \cmidrule(lr){6-8} \cmidrule(lr){9-11} \cmidrule(lr){12-14} \cmidrule(lr){15-15}
& & & \textbf{Cora} & \textbf{Photo} & \textbf{Physics} & \textbf{HM} & \textbf{Flickr} & \textbf{ArXiv} & \textbf{Reddit} & \textbf{MAG} & \textbf{Pokec} & \textbf{Products} & \textbf{WebTop.} & \textbf{Papers.} \\
\midrule

\multirow{4}{*}{\rotatebox{90}{Traditional}} 
& \multirow{2}{*}{KMeans}   & NMI & \res{13.89}{4.46} & \res{31.70}{0.49} & \res{52.00}{0.13} & \res{10.17}{0.11} & \res{1.21}{0.07} & \res{22.59}{0.03} & \res{11.08}{0.14} & \res{28.32}{0.06} & \res{1.41}{0.00} & \res{29.24}{0.11} & \res{2.45}{0.05} & \res{37.77}{0.07} \\
&                          & Mod & \res{18.73}{3.05} & \res{16.85}{0.40} & \res{51.56}{0.07} & \res{2.84}{0.05} & \res{0.90}{0.08} & \res{13.23}{0.31} & \res{4.48}{0.15} & \res{14.53}{0.06} & \res{0.89}{0.04} & \res{28.88}{0.28} & \res{0.56}{0.19} & \res{15.79}{0.09} \\
\cmidrule{2-15}
& \multirow{2}{*}{Node2Vec} & NMI & \res{44.95}{1.49} & \res{66.02}{1.05} & \res{54.94}{0.01} & \res{7.60}{0.13} & \res{5.59}{0.00} & \res{38.93}{0.15} & \res{79.23}{0.38} & \res{37.73}{0.03} & \second{31.92}{0.02} & \res{50.97}{0.20} & \res{5.80}{0.09} & \res{51.29}{0.05} \\
&                          & Mod & \res{71.52}{0.15} & \best{71.65}{0.14} & \res{59.19}{0.00} & \res{-5.89}{0.08} & \res{41.93}{0.01} & \res{55.48}{0.56} & \res{61.41}{0.58} & \res{45.08}{0.26} & \res{30.44}{0.16} & \res{77.17}{0.66} & \res{3.24}{0.12} & \res{42.23}{0.48} \\
\midrule

\multirow{6}{*}{\rotatebox{90}{Non-Parametric}} 
& \multirow{2}{*}{SSGC}     & NMI & \res{51.85}{1.05} & \res{70.75}{1.39} & \res{64.49}{4.20} & \res{12.15}{0.11} & \res{4.48}{0.19} & \res{46.12}{0.16} & \res{51.61}{0.36} & \second{41.52}{0.04} & \res{4.33}{0.01} & \res{52.04}{0.17} & \res{3.92}{0.11} & -- \\
&                          & Mod & \res{72.60}{0.35} & \res{70.13}{1.91} & \res{57.59}{4.45} & \res{4.85}{0.03} & \res{20.86}{0.30} & \res{60.21}{0.12} & \res{33.83}{1.19} & \res{64.64}{0.07} & \res{16.94}{0.41} & \res{80.58}{0.16} & \res{17.82}{0.79} & -- \\
\cmidrule{2-15}
& \multirow{2}{*}{SAGSC}    & NMI & \res{44.35}{0.06} & \res{58.40}{0.02} & \res{55.95}{0.04} & \res{12.47}{0.13} & \res{6.86}{0.00} & \res{43.27}{0.19} & \second{80.02}{0.38} & \res{40.15}{0.03} & \best{38.33}{0.03} & \res{51.78}{0.24} & \res{9.11}{0.26} & -- \\
&                          & Mod & \res{55.76}{0.05} & \res{65.11}{0.01} & \res{45.40}{0.01} & \res{-2.96}{0.39} & \res{24.73}{0.03} & \res{53.04}{2.24} & \second{68.43}{0.63} & \second{70.88}{0.31} & \best{50.01}{0.27} & \best{84.56}{0.17} & \res{16.81}{1.17} & -- \\
\cmidrule{2-15}
& \multirow{2}{*}{MS2CAG}   & NMI & \res{53.64}{0.64} & \second{72.49}{0.79} & \second{72.45}{0.03} & \res{9.75}{0.36} & \res{7.32}{0.02} & \res{44.20}{0.18} & \res{72.85}{0.53} & \res{40.39}{0.07} & \res{2.77}{0.05} & \res{50.71}{0.31} & \res{7.77}{0.08} & -- \\
&                          & Mod & \second{74.67}{0.15} & \res{71.08}{0.05} & \res{49.60}{0.02} & \res{2.50}{0.10} & \second{44.40}{0.03} & \second{64.06}{0.27} & \res{48.58}{1.89} & \res{63.42}{0.41} & \res{3.54}{0.17} & \second{83.80}{0.14} & \best{35.33}{0.46} & -- \\
\midrule

\multirow{12}{*}{\rotatebox{90}{Deep Decoupled}} 
& \multirow{2}{*}{GAE}      & NMI & \res{50.09}{0.07} & \res{61.36}{0.14} & \res{68.15}{0.03} & \second{13.58}{0.12} & \res{4.07}{0.05} & \res{40.86}{0.15} & \res{45.90}{0.27} & \res{39.30}{0.03} & \res{3.79}{0.04} & \res{42.55}{0.09} & \res{3.89}{0.14} & \res{42.84}{0.02} \\
&                          & Mod & \res{72.39}{0.11} & \res{66.94}{0.17} & \res{49.97}{0.02} & \res{3.49}{0.18} & \res{21.66}{0.53} & \res{58.22}{0.26} & \res{36.96}{0.35} & \res{54.26}{0.22} & \res{11.66}{0.07} & \res{62.17}{0.17} & \res{5.24}{0.24} & \res{25.79}{0.05} \\
\cmidrule{2-15}
& \multirow{2}{*}{DGI}      & NMI & \res{56.22}{0.87} & \res{67.77}{0.54} & \res{74.39}{0.35} & \res{11.67}{0.06} & \res{6.88}{0.01} & \res{42.29}{0.08} & \res{72.93}{0.14} & \res{39.39}{0.03} & \res{4.20}{0.01} & \res{41.28}{0.13} & \res{5.91}{0.02} & \res{49.28}{0.07} \\
&                          & Mod & \res{71.49}{0.28} & \res{64.99}{0.37} & \res{48.44}{0.09} & \res{5.53}{0.10} & \res{6.81}{0.25} & \res{54.30}{0.50} & \res{62.19}{0.27} & \res{56.43}{0.48} & \res{6.65}{0.04} & \res{61.54}{0.25} & \res{-1.07}{0.44} & \res{32.21}{0.20} \\
\cmidrule{2-15}
& \multirow{2}{*}{CCASSG}   & NMI & \res{58.74}{0.87} & \res{64.54}{3.23} & \res{70.94}{0.04} & \res{11.93}{0.15} & \res{4.67}{0.01} & \res{44.69}{0.04} & \res{49.63}{0.07} & \res{40.40}{0.02} & \res{1.35}{0.01} & \res{50.89}{0.31} & \res{3.79}{0.18} & \best{53.82}{0.05} \\
&                          & Mod & \res{73.46}{0.42} & \res{68.54}{1.69} & \res{49.12}{0.03} & \res{6.48}{0.76} & \res{-4.17}{0.46} & \res{49.95}{0.16} & \res{37.92}{0.34} & \res{57.57}{0.16} & \res{0.84}{0.00} & \res{74.64}{0.17} & \res{0.76}{0.71} & \best{50.06}{0.28} \\
\cmidrule{2-15}
& \multirow{2}{*}{S3GC}     & NMI & \res{55.45}{1.12} & \res{68.27}{2.22} & \res{70.87}{0.05} & \res{11.57}{0.08} & \second{7.84}{0.00} & \second{47.11}{0.12} & \best{83.45}{0.39} & \res{39.80}{0.03} & \res{6.04}{0.00} & \second{53.43}{0.16} & \res{7.83}{0.02} & \res{44.18}{0.04} \\
&                          & Mod & \res{74.49}{0.07} & \res{71.04}{0.07} & \res{49.72}{0.01} & \second{11.14}{0.30} & \res{39.96}{0.01} & \res{57.98}{0.39} & \best{70.94}{1.25} & \res{49.45}{0.26} & \res{13.53}{0.15} & \res{73.93}{0.33} & \res{29.06}{0.26} & \res{37.52}{0.21} \\
\cmidrule{2-15}
& \multirow{2}{*}{NS4GC}    & NMI & \best{59.40}{0.48} & \best{72.62}{0.79} & \best{75.38}{0.08} & \best{15.28}{0.17} & \res{6.19}{0.01} & \best{48.39}{0.24} & \res{56.71}{0.05} & \best{41.64}{0.01} & \res{7.19}{0.04} & \best{54.63}{0.14} & \best{10.05}{0.04} & \res{49.83}{0.04} \\
&                          & Mod & \best{75.08}{0.25} & \second{71.40}{0.01} & \res{48.94}{0.01} & \res{5.11}{0.19} & \res{-5.15}{0.05} & \res{62.37}{0.27} & \res{41.98}{0.14} & \res{62.80}{0.21} & \res{15.98}{0.09} & \res{77.89}{0.18} & \res{24.51}{0.06} & \res{42.44}{0.55} \\
\cmidrule{2-15}
& \multirow{2}{*}{MAGI}     & NMI & \second{58.94}{0.41} & \res{68.65}{0.16} & \res{66.08}{0.13} & \res{11.24}{0.46} & \res{6.31}{0.16} & \res{46.53}{0.16} & \res{72.53}{0.19} & \res{41.34}{0.02} & \res{8.76}{0.02} & \res{44.58}{0.14} & \second{9.27}{0.20} & \second{53.06}{0.04} \\
&                          & Mod & \res{71.19}{0.06} & \res{68.69}{0.09} & \res{60.53}{0.19} & \res{-4.15}{0.18} & \best{47.35}{0.08} & \res{61.71}{0.23} & \res{55.80}{0.43} & \res{64.73}{0.18} & \res{19.64}{0.07} & \res{68.51}{0.03} & \second{34.96}{1.23} & \second{47.63}{0.61} \\
\midrule

\multirow{10}{*}{\rotatebox{90}{Deep Joint}} 
& \multirow{2}{*}{DAEGC}    & NMI & \res{46.88}{1.71} & \res{63.60}{0.02} & \res{57.04}{0.03} & \res{11.46}{0.09} & \res{4.23}{0.03} & \res{39.81}{0.29} & \res{40.84}{0.76} & \res{29.00}{2.17} & \res{3.85}{0.21} & \res{9.18}{3.05} & \res{4.79}{0.29} & \res{28.59}{0.06} \\
&                          & Mod & \res{73.45}{0.62} & \res{65.34}{0.01} & \second{61.95}{0.02} & \res{4.40}{0.21} & \res{19.45}{0.09} & \res{62.65}{0.18} & \res{49.83}{1.08} & \res{61.72}{1.31} & \res{10.26}{0.31} & \res{35.59}{3.73} & \res{27.16}{1.13} & \res{41.13}{0.57} \\
\cmidrule{2-15}
& \multirow{2}{*}{DinkNet}  & NMI & \res{55.56}{0.18} & \res{64.74}{0.06} & \res{57.34}{0.06} & \res{11.26}{0.06} & \res{6.67}{0.05} & \res{37.24}{0.57} & \res{54.95}{0.24} & \res{37.12}{0.05} & \res{3.91}{0.01} & \res{38.43}{0.18} & \res{5.47}{0.04} & \res{45.86}{0.04} \\
&                          & Mod & \res{72.26}{0.16} & \res{64.98}{0.21} & \res{60.92}{0.06} & \res{5.93}{0.22} & \res{4.37}{0.52} & \res{31.87}{0.36} & \res{42.73}{0.38} & \res{34.56}{0.27} & \res{12.47}{0.34} & \res{59.75}{0.25} & \res{-2.34}{0.28} & \res{32.51}{0.43} \\
\cmidrule{2-15}
& \multirow{2}{*}{MinCut}   & NMI & \res{40.80}{1.83} & \res{62.34}{2.39} & \res{56.94}{2.06} & \res{7.42}{0.37} & \res{7.59}{0.19} & \res{39.00}{1.02} & \res{48.85}{3.58} & \res{38.21}{0.12} & \res{5.78}{0.18} & \res{35.80}{0.71} & \res{6.07}{0.52} & -- \\
&                          & Mod & \res{72.99}{1.28} & \res{67.00}{1.75} & \res{61.47}{0.20} & \res{4.25}{2.35} & \res{29.50}{4.82} & \res{61.80}{1.08} & \res{42.38}{1.55} & \res{68.97}{0.25} & \res{14.55}{0.43} & \res{73.38}{0.30} & \res{24.86}{1.25} & -- \\
\cmidrule{2-15}
& \multirow{2}{*}{DMoN}     & NMI & \res{43.84}{2.29} & \res{62.74}{2.67} & \res{58.30}{0.37} & \res{7.59}{0.92} & \best{7.89}{0.09} & \res{38.77}{0.28} & \res{50.66}{0.38} & \res{38.30}{0.12} & \res{8.42}{0.08} & \res{34.80}{0.85} & \res{7.49}{0.37} & -- \\
&                          & Mod & \res{71.81}{1.40} & \res{70.32}{0.28} & \best{62.90}{0.11} & \best{12.54}{0.25} & \res{37.17}{0.74} & \res{61.14}{1.19} & \res{47.78}{1.70} & \res{69.58}{0.15} & \res{22.67}{0.28} & \res{68.35}{0.53} & \res{33.90}{0.23} & -- \\
\cmidrule{2-15}
& \multirow{2}{*}{Neuromap} & NMI & \res{46.98}{2.61} & \res{61.84}{2.34} & \res{56.53}{1.63} & \res{7.44}{0.17} & \res{7.77}{0.57} & \res{40.70}{0.80} & \res{47.81}{0.28} & \res{39.14}{0.20} & \res{8.53}{0.25} & \res{34.80}{1.46} & \res{7.19}{0.51} & -- \\
&                          & Mod & \res{74.02}{0.95} & \res{69.10}{1.96} & \res{57.35}{0.33} & \res{6.29}{0.17} & \res{34.35}{2.98} & \best{64.54}{0.66} & \res{41.11}{0.27} & \best{76.25}{0.15} & \second{31.71}{0.56} & \res{70.59}{0.43} & \res{31.96}{1.39} & -- \\
\bottomrule
\end{tabular}
\end{adjustbox}
\vspace{-2mm}
\end{table*}

%% file: tables/efficiency.tex
\begin{table*}[t]
\centering
\caption{\textbf{Efficiency profiling results for larger datasets on a single NVIDIA V100 GPU (32GB).}
Mem: Peak GPU Memory; Time: Total Training and Clustering Time.
Results are drawn from our companion benchmark~\cite{PyAGC}.}
\label{tab:efficiency}
\small
\resizebox{\textwidth}{!}{
\begin{tabular}{ll|cc|cc|cc|cc|cc|cc}
\toprule
\multirow{2}{*}{\textbf{Category}} & \multirow{2}{*}{\textbf{Method}} & \multicolumn{2}{c|}{\textbf{Reddit}} & \multicolumn{2}{c|}{\textbf{MAG}} & \multicolumn{2}{c|}{\textbf{Pokec}} & \multicolumn{2}{c|}{\textbf{Products}} & \multicolumn{2}{c|}{\textbf{WebTopic}} & \multicolumn{2}{c}{\textbf{Papers100M}} \\
\cmidrule(lr){3-4} \cmidrule(lr){5-6} \cmidrule(lr){7-8} \cmidrule(lr){9-10} \cmidrule(lr){11-12} \cmidrule(lr){13-14}
 & & Mem(GB) & Time(m) & Mem(GB) & Time(m) & Mem(GB) & Time(m) & Mem(GB) & Time(m) & Mem(GB) & Time(m) & Mem(GB) & Time(h) \\
\midrule
\multirow{2}{*}{Traditional} & KMeans & - & 0.27 & - & 6.59 & - & 3.09 & - & 5.87 & - & 3.98 & - & 0.05 \\
 & Node2Vec & 0.70 & 22.67 & 1.90 & 77.58 & 4.04 & 236.53 & 15.05 & 64.94 & 11.80 & 75.92 & 7.20 & 12.38 \\
\midrule
\multirow{3}{*}{\shortstack[l]{Non-\\Parametric}} & SSGC & - & 5.36 & - & 6.51 & - & 4.43 & - & 8.05 & - & 10.38 & - & - \\
 & SAGSC & - & 5.99 & - & 23.35 & - & 4.72 & - & 5.29 & - & 2.77 & - & - \\
 & MS2CAG & - & 0.16 & - & 15.10 & - & 13.68 & - & 1.38 & - & 16.81 & - & - \\
\midrule
\multirow{5}{*}{\shortstack[l]{Deep\\Decoupled}} & GAE & 14.18 & 30.58 & 10.16 & 26.54 & 27.97 & 10.16 & 13.33 & 26.12 & 1.63 & 6.18 & 7.58 & 0.67 \\
 & DGI & 24.16 & 1.93 & 22.46 & 10.54 & 30.90 & 10.37 & 29.52 & 50.32 & 8.96 & 6.16 & 30.89 & 2.42 \\
 & CCASSG & 27.50 & 1.21 & 28.64 & 31.14 & 30.75 & 117.43 & 30.30 & 45.16 & 6.16 & 122.30 & 31.09 & 1.29 \\
 & S3GC & 2.82 & 32.38 & 5.12 & 41.78 & 9.01 & 71.17 & 29.46 & 29.68 & 18.98 & 145.98 & 15.85 & 5.85 \\
 & NS4GC & 10.90 & 12.84 & 17.71 & 41.04 & 30.12 & 55.88 & 24.10 & 37.24 & 8.95 & 30.75 & 13.97 & 1.24 \\
 & MAGI & 20.41 & 56.98 & 30.77 & 86.99 & 21.44 & 176.43 & 31.08 & 375.99 & 5.23 & 115.36 & 29.28 & 6.04 \\
\midrule
\multirow{5}{*}{\shortstack[l]{Deep\\Joint}} & DAEGC & 31.20 & 14.28 & 26.18 & 27.52 & 30.90 & 14.28 & 15.37 & 37.28 & 11.71 & 11.11 & 15.96 & 1.40 \\
 & DinkNet & 24.16 & 4.49 & 23.66 & 5.46 & 24.84 & 6.18 & 30.63 & 85.27 & 29.98 & 4.27 & 31.11 & 5.13 \\
 & MinCut & 18.56 & 6.73 & 18.16 & 16.01 & 19.32 & 24.15 & 30.14 & 32.55 & 28.27 & 8.37 & - & - \\
 & DMoN & 18.56 & 1.22 & 18.16 & 15.46 & 19.32 & 22.91 & 30.14 & 22.88 & 28.27 & 6.75 & - & - \\
 & Neuromap & 18.57 & 6.73 & 18.16 & 14.56 & 19.32 & 23.07 & 24.50 & 23.00 & 28.27 & 4.97 & - & - \\
\bottomrule
\end{tabular}}
\vspace{-2mm}
\end{table*}